\documentclass[runningheads]{llncs}

\usepackage{eccv}

\usepackage{eccvabbrv}

\usepackage{graphicx}
\usepackage{booktabs}

\usepackage[accsupp]{axessibility}  

\usepackage{multirow}

\usepackage{hyperref}

\usepackage{orcidlink}
\usepackage{marvosym}
\usepackage{tabu}

\begin{document}

\title{NGP-RT: Fusing Multi-Level Hash Features with Lightweight Attention for Real-Time Novel View Synthesis} 

\titlerunning{NGP-RT}


\author{
Yubin Hu\inst{1}\textsuperscript{*}\textsuperscript{$\dag$} \and
Xiaoyang Guo\inst{2}\textsuperscript{*} \and
Yang Xiao \inst{3}
\and \\ 
Jingwei Huang \inst{4}
\and  
Yong-Jin Liu \inst{1}\textsuperscript{\Letter}}

\authorrunning{Y. Hu et al.}

\institute{
BNRist, Department of Computer Science and Technology, Tsinghua University
\and
\mbox{Horizon Robotics \qquad \and Huawei Technologies \qquad \and Game AI Center, Tencent Games}}

\maketitle

\def\thefootnote{*}
\footnotetext{These authors contributed equally to this work.}
\def\thefootnote{\dag}
\footnotetext{The work was done while the author was an intern at Huawei.}
\def\thefootnote{\Letter}
\footnotetext{Corresponding author.}
\def\thefootnote{\arabic{footnote}}

\begin{abstract}
This paper presents NGP-RT, a novel approach for enhancing the rendering speed of Instant-NGP to achieve real-time novel view synthesis.
As a classic NeRF-based method, Instant-NGP stores implicit features in multi-level grids or hash tables and applies a shallow MLP to convert the implicit features into explicit colors and densities. Although it achieves fast training speed, there is still a lot of room for improvement in its rendering speed due to the per-point MLP executions for implicit multi-level feature aggregation, especially for real-time applications. To address this challenge, our proposed NGP-RT explicitly stores colors and densities as hash features, and leverages a lightweight attention mechanism to disambiguate the hash collisions instead of using computationally intensive MLP. 
At the rendering stage, NGP-RT incorporates a pre-computed occupancy distance grid into the ray marching strategy to inform the distance to the nearest occupied voxel, thereby reducing the number of marching points and global memory access. Experimental results show that on the challenging Mip-NeRF 360 dataset, NGP-RT achieves better rendering quality than previous NeRF-based methods, achieving \textbf{108 fps} at \textbf{1080p} resolution on a single Nvidia RTX 3090 GPU. Our approach is promising for NeRF-based real-time applications that require efficient and high-quality rendering.
\end{abstract}

\section{Introduction}
\label{sec:intro}

\begin{figure*}[tb]
  \centering 
  \includegraphics[width=1.0\linewidth]{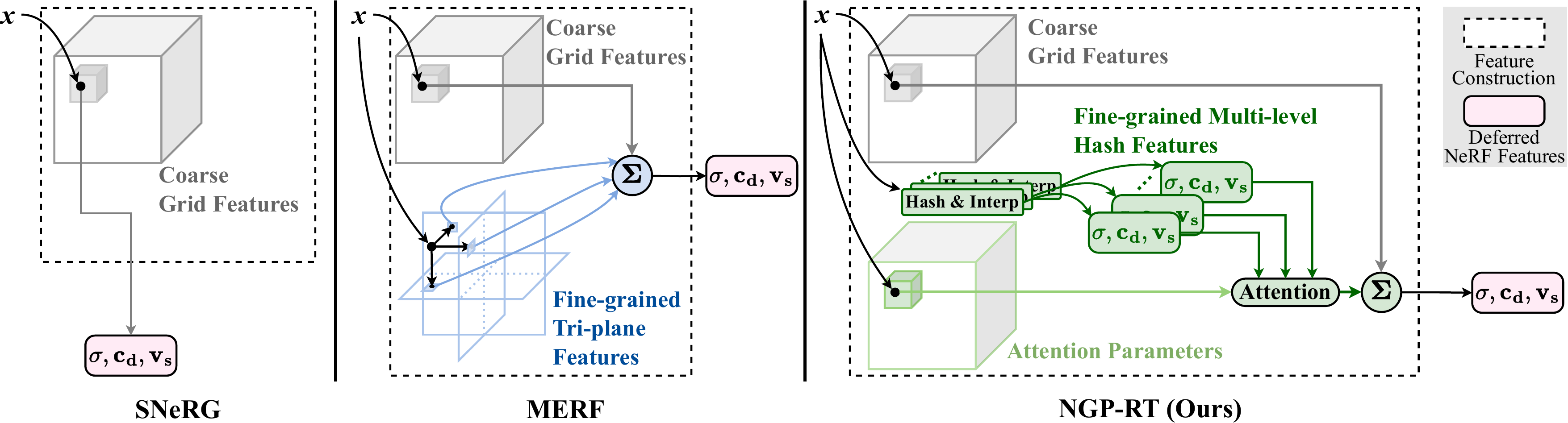}
  \caption{%
    Comparisons of the feature construction methods in SNERG~\cite{snerg}, MERF~\cite{reiser2023merf}, and our NGP-RT. We utilize a lightweight attention mechanism to efficiently aggregate the multi-level explicit hash features, which sufficiently exploit the expressing power of NGP-style features under the deferred NeRF architecture for fast rendering. 
    %
  }
  \label{fig:snerg-merf-ours}
\end{figure*}

Novel view synthesis is a long-standing problem in computer vision. 
The emergence of neural radiance fields~\cite{nerf,nerf++,mipnerf,mipnerf360} has led to state-of-the-art photorealistic view synthesis using the neural volumetric scene representation. 
However, in these pioneering works, the original MLP parameterization exhibits high computational complexity.
To address this limitation, recent works have proposed replacing or augmenting the MLPs with feature grids~\cite{dvgo,plenoxels}, tri-planes~\cite{tri-miprf}, multi-plane images~\cite{he2023mmpi,kohler2023fmpi} or tensorial vectors~\cite{tensorf,strivec}. 

Among them, Instant-NGP \cite{muller2022instant} stands out as a promising technique that utilizes multi-level implicit feature grids to represent the radiance field and employs a shallow MLP to decode implicit features into explicit colors and densities.
By storing high-resolution implicit features in limited-length hash tables, Instant-NGP achieves high-quality novel view synthesis after a short optimization period.
Despite its advantages in training speed and rendering quality, Instant-NGP cannot meet the real-time rendering requirements for large scenes. 
The shallow MLP for feature aggregation of each querying point slows down the rendering speed, even though accessing the multi-level grid features is relatively fast.
However, simply reducing the MLP parameters or removing the MLP would result in severe rendering quality degradation.

Recent works, such as SNeRG \cite{snerg} and MERF \cite{reiser2023merf}, proposed a deferred NeRF architecture to remove the per-point MLP.
They directly store explicit colors and densities instead of implicit features, and only execute MLP once for each casting ray, which highly accelerates the rendering process. 
However, their treatments of the explicit features show limited expressing power and are unsuitable for multi-level features of Instant-NGP.
As shown in Figure \ref{fig:snerg-merf-ours}, the low-resolution sparse grid of SNeRG fails to cover the detailed features at fine-grained resolution levels of Instant-NGP. At the same time, the simple sum aggregations in MERF are not flexible enough to compensate for hash collisions in high-resolution NGP features. 
Applying these feature construction methods directly to the multi-level features in Instant-NGP may lead to a compromise in their representation ability and result in a degradation of rendering quality.

To inherit the rendering efficiency of the deferred NeRF architecture and preserve the strong representation ability of Instant-NGP, we propose NGP-RT, a novel method that utilizes a lightweight attention mechanism to render high-fidelity novel views efficiently. 
The attention mechanism of NGP-RT employs simple yet effective weighted sum operations with learnable attention parameters that adaptively prioritize the explicit multi-level hash features.
Considering different modalities in the deferred NeRF features, we assign two attention parameters 
to each resolution level: one for the density value and the other for the color features.
Moreover, to mitigate the hash collisions that map different positions to the same entry point of the hash table, we further design the attention parameters to be spatially distinct for hash features extracted from different positions.
With these independent learnable parameters, positions mapped to the same entry point can flexibly adjust their relative importance to optimize the multi-view rendering results.

When rendering, we store the pre-computed attention parameters in a low-resolution grid. 
For each queried point on the casting ray, NGP-RT samples the attention parameters and applies channel-wise weighted sum operations to the multi-level explicit features.
The lightweight aggregations reduce over 90\% of the multiply-accumulate (MAC) operations compared to the shallow MLP of Instant-NGP, 
enabling the high-speed rendering process of NGP-RT.

With our design that alleviates the computational burden, global memory access has gradually become a bottleneck in the rendering process. 
Since the data access of point-wise features is necessary and challenging to shorten, we focus on reducing another source of global memory access: the occupancy check that accesses the multi-level occupancy grids.
With access to the occupancy grids saved in global memory, existing methods~\cite{muller2022instant,reiser2023merf} perform occupancy checks for every marching point to skip empty spaces in the ray marching strategies. 
However, the typical voxel-by-voxel ray marching method results in a large number of marching points and occupancy checks, congesting the memory bandwidth. 
To address this issue, we incorporate a pre-computed occupancy distance grid into the ray marching strategy, which informs the distance to the nearest occupied voxel and allows for fewer marching steps. 
As a result, our method can reduce global memory access without sacrificing visual quality.

We summarize our contributions as follows: 
(1) We design a lightweight attention mechanism with learnable parameters to efficiently aggregate the multi-level hash features. (2) We incorporate the occupancy distance grid to reduce global memory access of occupancy checks and further accelerate the rendering process. (3) By leveraging the explicit multi-level hash features, we develop a real-time NeRF method that renders high-quality 1080p images at over 100 fps.

\section{Related Work}

In recent years, the neural radiance field~\cite{nerf} has attracted significant attention for its capability in 3D geometry modeling and high-fidelity novel view synthesis. 
Previous work has explored different variants of NeRFs~\cite{mipnerf,relu-fields,hollownerf,f2-nerf} to improve the rendering quality and training speed. 
However, achieving fast rendering speed for large-scale scenes with high quality at high resolution still remains challenging.
In this section, we mainly review the existing techniques for rendering acceleration of NeRF-based methods, which can be divided into two main categories with different motivations.

One category of approaches employs rasterization on geometry proxies to alleviate the complexities associated with volumetric rendering, such as point clouds~\cite{npbg++}, meshes~\cite{mobilenerf,bakedsdf,wan2023duplex,tang2023delicate,re-rend,svs} and planes~\cite{lin2022neurmips}. 
MobileNeRF~\cite{mobilenerf} represents radiance fields using a set of polygons with neural textures, and employs a traditional rasterization pipeline to achieve interactive frame rates.  
Wan~\etal~\cite{wan2023duplex} proposes representing scenes using a two-layer duplex mesh with neural features.
BakedSDF~\cite{bakedsdf} optimizes a hybrid neural volume-surface scene representation baked into a high-quality triangle mesh for rendering. 
Although these methods exhibit excellent real-time performance, they produce suboptimal rendering results at object edges and incorrectly reconstructed surfaces, especially for large-scale 360-degree scenes.
In contrast to NeRF-based methods based on volumetric neural representations, 3D Gaussians~\cite{gaussian} have emerged as a promising new representation for real-time rendering, but require high storage consumption when rendering large-scale scenes with a massive number of 3D Gaussians.
Moreover, it can be limited by its need for sparse point initialization and can exhibit spike-shaped artifacts when viewed from specific angles.

The other category of methods preserves the volumetric rendering scheme to maintain superior rendering quality, which can be further divided into several subcategories. 
\textbf{1)} Firstly, there are methods that focus on reducing the number of sampled points per ray with early termination~\cite{terminerf} and adaptive sampling~\cite{adanerf,donerf}, or decreasing the number of pixels to be rendered with the render-then-upsample strategy~\cite{4k-nerf,steernerf,wu2022scalable}.
Most of these techniques are orthogonal to the discussion of this paper and can be applied to our method for further speed improvements. 
\textbf{2)} Another subset of methods aims to propose more lightweight alternatives for the per-point deep MLPs, such as multiple small MLPs~\cite{kilonerf,wu2022scalable} and the MIMO MLP architecture~\cite{mimonerf}. 
Neural light fields~\cite{r2l,realtime-lightfield} encode the light field into a neural network and require only a single network forward pass for the rendering of each pixel. Although these methods improve the rendering speed of NeRF to some extent, the presence of the remaining MLP still prevents them from rendering fast at high resolutions.
\textbf{3)} The third typical solution involves explicitly representing the radiance fields and completely removing the MLP for rendering.
Many works store the explicit color and density values in the form of voxel grids~\cite{dvgo,plenoxels,deng2023compressing}, octrees~\cite{plenoctree}, VDB structures~\cite{plenvdb}, and NerfTrees~\cite{efficientnerf}. 
In contrast to directly caching the colors and densities, FastNeRF~\cite{fastnerf} and its variants~\cite{ccnerf,squeezenerf} compactly cache the factorized radiance fields and obtain the color values with efficient inner products. 
While these methods have demonstrated real-time performance on small objects, they tend to consume excessive storage and GPU memory for high-quality renderings of larger scenes.

Within this context, deferred neural rendering techniques~\cite{snerg,reiser2023merf,hu2023multiscale,zhang2022digging} combine the storage of explicit features with a tiny view-dependent MLP. 
SNeRG~\cite{snerg} proposes the deferred rendering method, allowing to evaluate the MLP only once for each pixel. However, the voxel representation encounters challenges in efficiently representing scenes at high resolutions. MERF~\cite{reiser2023merf} addresses this limitation by integrating high-frequency features with high-resolution 2D feature planes. However, its expressive power is still limited compared to multi-level hash features~\cite{muller2022instant}. 
Our goal is to enhance the representation capability of existing deferred NeRFs for large-scale 360-degree scenes by incorporating powerful multi-level hash features.

\section{Preliminaries}

Our method is built upon the original NeRF method \cite{nerf}, Instant-NGP \cite{muller2022instant}, and the deferred NeRF architecture. We briefly introduce them below. 

The original NeRF represents a 3D scene with 3D fields of densities $\sigma$ and colors ${\bf c}$, and models both fields as continuous functions of the 3D point coordinate ${\bf x} \in \mathbb{R}^3$ and view direction ${\bf d} \in \mathbb{S}^2$.
Specifically, NeRF parameterizes the functions using an MLP, which can be formulated as follows:  
\begin{equation}
\label{eq:nerf}
    \sigma, {\bf c} = {\rm MLP}_\Theta({\bf x}, {\bf d}). 
\end{equation}
To render the color ${\bf C}$ of a ray emitted from point ${\bf o}$ in the direction ${\bf d}$, NeRF queries the MLP at 3D points sampled along the ray, ${\bf x}_i = {\bf o} + t_i {\bf d}$, and composites the resulting densities $\{\sigma_i\}$ and colors $\{{\bf c}_i\}$ according to the approximated volume rendering integral, as discussed by Max \cite{max1995optical},
\begin{equation}
\begin{gathered}
    {\bf C} = \sum\limits_{i=1}^{N} \alpha_i T_i {\bf c}_i, \quad  
    T_i = \prod\limits_{j=1}^{i-1} (1-\alpha_j), \quad  \alpha_i = 1 - e^{-\sigma_i\delta_i},
\end{gathered}
\label{eq:volume_render}
\end{equation}
where $T_i$ and $\alpha_i$ denote the transmittance and alpha values of sample $i$, and $\delta_i = t_{i+1} - t_i$ is the distance between two adjacent samples. 

Similarly, Instant-NGP represents the 3D scene with color and density fields but is parameterized by implicit multi-level grid features along with a tiny MLP. This parameterization can be formulated as:
\begin{equation}
    \sigma, {\bf c} = {\rm MLP}_\theta([{\bf F}_{1, {\bf x}}, {\bf F}_{2, {\bf x}}, \dots {\bf F}_{K,{\bf x}}], {\bf d}),
\end{equation}
where ${\rm MLP}_\theta$ is shallower than ${\rm MLP}_\Theta$ in Eq.(\ref{eq:nerf}), ${\bf F}_{k,{\bf x}}$ denotes the level-$k$ feature of the 3D position ${\bf x}$ extracted from the corresponding grid or hash table, and $[\cdot]$ denotes the feature concatenation operation. The success of Instant-NGP demonstrates the strong expressing power of multi-level hash features, which motivates our utilization of the NGP-style features in real-time NeRF rendering.

To further reduce the number of MLP evaluations from $N$ to $1$ for each ray, SNeRG~\cite{snerg} utilizes a deferred shading model, in which the 3D scene is represented by the 3D fields of densities $\sigma$, diffuse colors ${\bf c_d}$, and specular feature vector ${\bf v_s}$. 
At the training time, these fields are parameterized by a deep MLP with the 3D coordinate ${\bf x}$ as input:
\begin{equation}
\label{eq:deferred}
    \sigma, {\bf c_d}, {\bf v_s} = {\rm MLP}_\Phi({\bf x}). 
\end{equation}
After the training stage, these optimized fields are baked into a coarse grid that directly stores $[\sigma, {\bf c_d}, {\bf v_s} ]$ as explicit deferred NeRF features, such that the point-wise density $\sigma_i$, diffuse color ${\bf c}^i_{\bf d}$, and feature vector ${\bf v}^i_{\bf s}$ can be directly obtained from the feature grid without any MLP execution.
After the volumetric accumulation, SNeRG computes another view-dependent color term for each casting ray via a tiny MLP.
The entire deferred rendering process of SNeRG can be formulated as
\begin{equation}
\label{eq:deferred_render}
\begin{gathered}
    {\bf C_d} = \sum\limits_{i=1}^{N} \alpha_i T_i {\bf c}^i_{\bf d}, \quad {\bf F} = \sum\limits_{i=1}^{N} \alpha_i T_i {\bf v}^i_{\bf s}, 
    \\
    {\bf C} = {\bf C_d} + {\rm MLP}_\psi({\bf C_d}, {\bf F}, {\bf d}),
\end{gathered}
\end{equation}
where $T_i$ and $\alpha_i$ follow the notations in Eq. (\ref{eq:volume_render}), and ${\rm MLP}_\psi$ denotes the tiny MLP for computing view-dependent colors. Such a combination of deferred NeRF and feature baking makes the fast rendering of SNeRG possible.

\section{Method}

\subsection{NGP-RT Overview}
\label{sec: overview}

Given a set of calibrated multi-view images of a static scene, 
our goal is to build a NeRF model that can be rendered in real time with high quality based on the superior expressive power of NGP-style multi-level hash features. 
However, directly integrating NGP-style features into the deferred NeRF architecture significantly reduces the rendering speed, since the indispensable MLP aggregations for each sample point consume a lot of time.

The tiny MLP used for decoding the implicit features into explicit colors and densities plays a crucial role in alleviating the gradient average problem caused by hash collision.
Hidden neurons of the tiny MLP can learn an adaptive masking function that identifies regions dominated by low-frequency textures according to the low-level features, and assigns relatively lower importance to high-level features in these regions.
Such an implicit masking effect greatly reduces the redundant usage of high-resolution features and alleviates the collisions of 3D positions mapped to the same hash table entry. 
Therefore, when using explicit hash features without the subsequent MLP, the importance assigned for conflicted 3D positions cannot be flexibly adjusted according to the contents of 3D scenes, which severely limits the rendering performance.

\begin{figure*}[tbp]
  \centering 
  \includegraphics[width=1.0\linewidth]{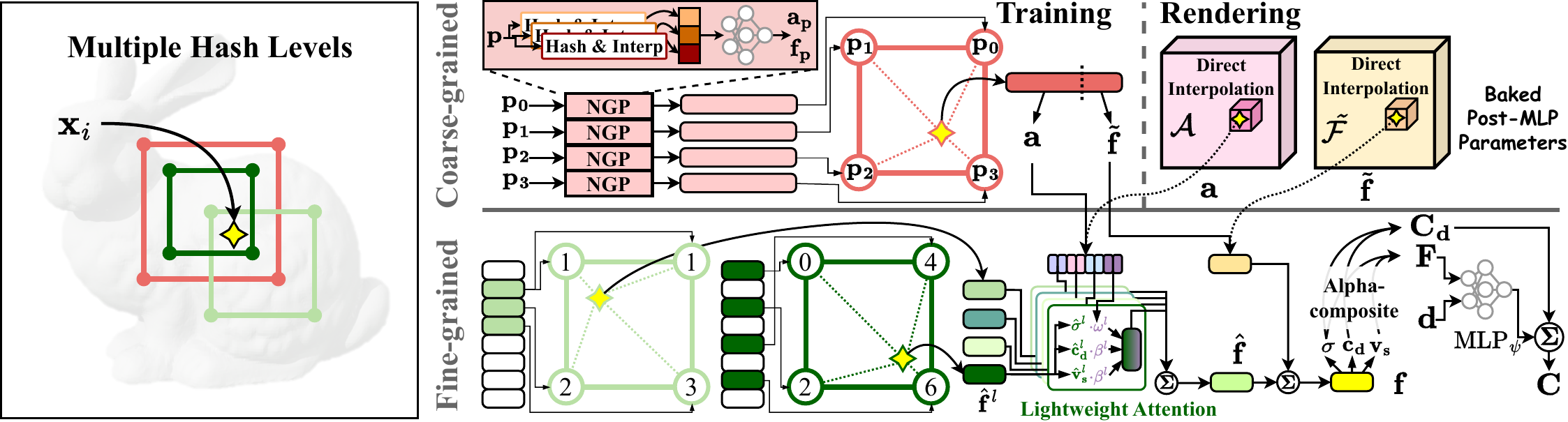}
  \caption{%
    2D illustration of the feature construction pipeline of NGP-RT. Following the practice in MERF \cite{reiser2023merf}, NGP-RT constructs the deferred NeRF feature ${\bf f}$ with a coarse-grained part $\Tilde{\bf f}$ and a fine-grained part $\hat{\bf f}$. At the training stage, we optimize $\Tilde{\bf f}$ and the attention parameters ${\bf a}$ with an auxiliary NGP model. At the inference stage, we bake them into the low-resolution grids $\Tilde{\mathcal{F}}$ and $\mathcal{A}$ for fast access and real-time rendering. The fine-grained features $\hat{\bf f}$ is fused from the high-resolution hash features fused by lightweight attention mechanism. NGP-RT employs the rendering process of deferred NeRF \cite{snerg} for volume rendering. We omit most of the superscript $i$ for simplicity.}
  \label{fig:overall}
  \vspace{-5mm}
\end{figure*}

To address the above challenge, we propose NGP-RT, a real-time NeRF method that incorporates a lightweight attention mechanism to assign spatially varying importance to high-level hash features without sacrificing the rendering speed.
As illustrated in Figure \ref{fig:overall}, NGP-RT divides the multi-level features into the coarse-grained and fine-grained parts. 
Given a 3D sample point ${\bf x}_i$, we construct the deferred NeRF feature ${\bf f}_i = [{\sigma}_i, {{\bf c}}^i_{\bf d}, {\bf v}^i_{\bf s}]$ at this position by summing the coarse-grained feature $\Tilde{\bf f}_i = 
 [\Tilde{\sigma}_i, \Tilde{{\bf c}}^i_{\bf d}, \Tilde{\bf v}^i_{\bf s}]$ and the fine-grained feature  $\hat{\bf f}_i = [\hat{\sigma}_i, \hat{{\bf c}}^i_{\bf d}, \hat{\bf v}^i_{\bf s}]$.
After feature extraction, the resulting per-point features $\{{\bf f}_i
| i=1, \dots, N \}$ are then fed into the deferred NeRF volume rendering process to compute the RGB color values ${\bf C}$ using Eq. (\ref{eq:deferred_render}).
In the following parts, we omit the superscript $i$ for simplicity.

For the coarse-grained level, we follow MERF~\cite{reiser2023merf} to optimize $\Tilde{\bf f}$ with an auxiliary NGP model at the training stage.
Subsequently, we bake the coarse-grained features into a sparse 3D voxel grid $\Tilde{\mathcal{F}}$ with a resolution $L_C$ for fast feature access.
As for the fine-grained feature $\hat{\bf f}$, we model it as the aggregation of multi-level hash features from $L$ high resolutions.
We design a lightweight attention mechanism to aggregate the fine-level features with channel-wise weighted sum operations according to the learnable attention parameters. 
The attention parameters ${\bf a}$ vary spatially and are decoded from the low-level branch together with the coarse-grained feature $\Tilde{\bf f}$, which effectively models the dependency of the masking function on 
coarse-grained contents.
We bake the attention parameters into a sparse grid $\mathcal{A}$ of resolution $L_C$ similar to $\Tilde{\mathcal{F}}$ for real-time rendering.

Overall, the feature construction of NGP-RT at the rendering stage can be formulated as 
\begin{equation}
\begin{gathered} 
    {\bf f} = \Tilde{\bf f} + \hat{\bf f},
    \\
    \Tilde{\bf f} = {\rm Interp}(\Tilde{\mathcal{F}}, {\bf x}), \quad {\bf a} = {\rm Interp}(\mathcal{A}, {\bf x}),
    \\
    \hat{\bf f} = {\rm Att}(\hat{\bf f}^1, \dots, \hat{\bf f}^L ; {\bf a}), 
\end{gathered}
\end{equation}
where ${\rm Interp}(\mathcal{Y}, {\bf x})$ denotes the function that accesses the feature at position ${\bf x}$ from feature grid $\mathcal{Y}$ using tri-linear interpolation, and $ {\rm Att}({\bf f}^1, \dots, {\bf f}^L ; {\bf a})$ denotes our lightweight attention mechanism that aggregates the multi-level features $\{{\bf f}^1, \dots, {\bf f}^L\}$ with the attention parameters ${\bf a}$.

At the training stage, we optimize ${\bf a}$ and  $\Tilde{\bf f}$ with an auxiliary NGP model, which aggregates auxiliary hash features into the attention parameters and coarse-grained features via a shallow MLP. 
To align with the direct feature interpolation at the rendering stage, we evaluate the auxiliary NGP model at grid corners of resolution $L_C$, and then conduct tri-linear interpolation to obtain ${\bf a}$ and $\Tilde{\bf f}$.
After training, we discard the auxiliary NGP model is discarded and directly obtain ${\bf a}$ and $\Tilde{\bf f}$ from the sparse voxel grids $\mathcal{\Tilde{F}}$ and $\mathcal{A}$, respectively.

\subsection{Feature Fusion with Lightweight Attention}

Our lightweight attention mechanism utilizes simple yet effective weighted sum operations with learnable parameters to aggregate features from different fine-grained resolution levels into $\hat{\bf f}$. 
We first design the attention parameters to be different for all fine-grained levels, such that they can help to balance the overload of different resolution levels and make features from different levels focus on different frequencies.
Furthermore, by imitating the implicit masking function inside the tiny MLP of Instant-NGP, we design the attention parameters to be spatially different by formalizing them as part of the decodings from the low-resolution branch.
Through the gradients back-propagated from the multi-view rendering loss, these parameters learn the appropriate importance of different spatial positions to alleviate the hash collisions and fully exploit the expressing power of hash features.
Considering that the explicit deferred feature $\hat{\bf f}^l$ consists of density $\hat{\sigma}^l$ and color features $[\hat{\bf c}^l_{\bf d}, \hat{\bf v}^l_{\bf s}]$, we utilize two attention parameters to model the spatial importance of the two modalities separately. 
Consequently, the attention parameters at position ${\bf x}$ can be decomposed to
\begin{equation}
    {\bf a} = [\omega^{1}, \beta^{1}, \dots,  \omega^{L}, \beta^{L}],
\end{equation}
where $\omega^{l}$ and $\beta^{l}$ denote attention parameters for $\hat{\sigma}^l$ and $[\hat{\bf c}^l_{\bf d}, \hat{\bf v}^l_{\bf s}]$ of the $l^{th}$ fine-grained resolution level, respectively.

Our attention operation multiplies these attention parameters to the corresponding features, formulated by
\begin{equation}
\begin{split}
    \hat{\bf f} & = {\rm Att}(\hat{\bf f}^1, \dots, \hat{\bf f}^L ; {\bf a}) \\
    & = \left[ \sum_{l=1}^{L} \omega^{l} \cdot \hat{\sigma}^l, \ \ \sum_{l=1}^{L} \beta^{l} \cdot \hat{\bf c}_{\bf d}^{l}, \ \ \sum_{l=1}^{L} \beta^{l} \cdot \hat{\bf v}_{\bf s}^{l} \right],
\end{split}
\end{equation}
where $\hat{\bf f}^l = [\hat{\sigma}^l, \hat{{\bf c}}^{l}_{\bf d}, \hat{\bf v}^{l}_{\bf s}]$ denotes the fine-grained features at the $l^{th}$ fine-grained resolution level.
Our lightweight attention involves only a few MAC operations for each sample point ${\bf x}$, and is fast enough to be directly applied to the rendering process.
Thus, we do not need to bake the aggregated high-resolution features at specific voxel grids, essentially preserving the details in the fine-level features.

The proposed lightweight attention mechanism effectively resolves explicit hash feature collisions with minimal computational cost, which is previously under-explored. This advance could also help other methods utilizing MLP for feature aggregation, such as the MLP-enhanced 3D Gaussians in 3D AIGC \cite{zou2023triplane} and human reconstruction \cite{kocabas2023hugs}.

\subsection{Ray Marching with Occupancy Distance}

\begin{figure}[htbp]
  \centering 
  \includegraphics[width=0.9\linewidth]{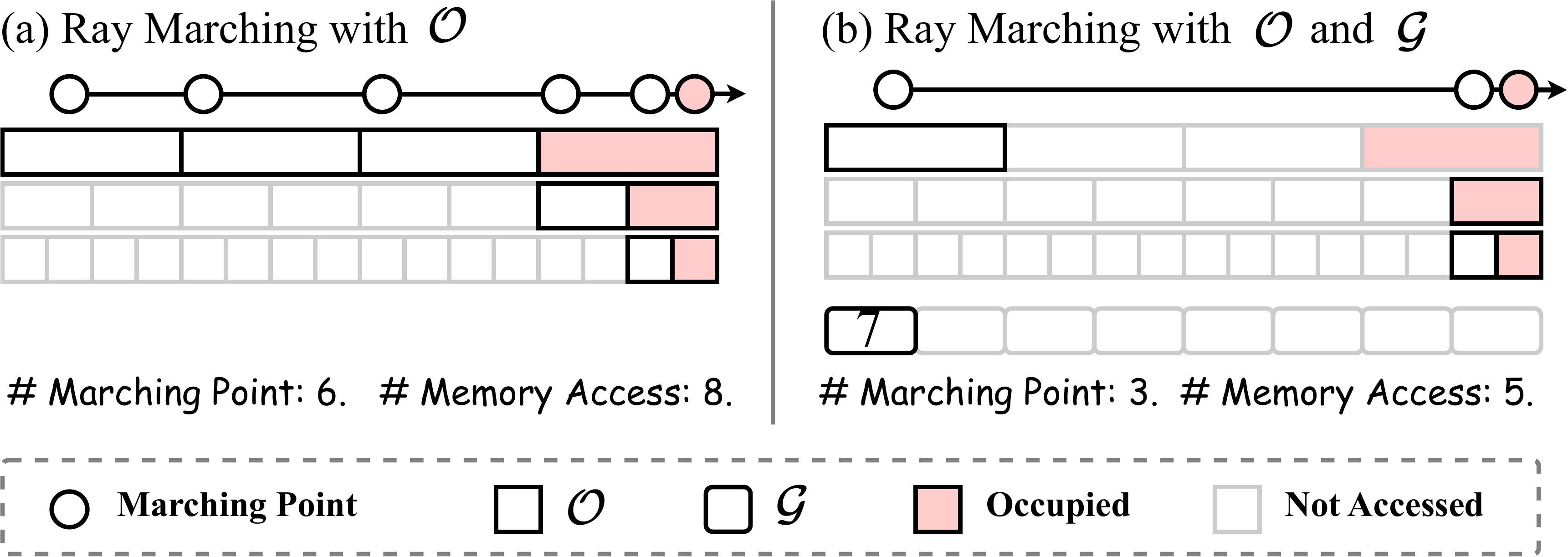}
  \caption{%
    Comparison between (a) the previous ray marching strategy with only multi-level occupancy grid $\mathcal{O}$ and (b) our strategy with $\mathcal{O}$ and occupancy distance grid $\mathcal{G}$.
  }
  \label{fig:occ}
\end{figure}

At the rendering stage, previous methods~\cite{muller2022instant,reiser2023merf} typically utilize the multi-level occupancy grids $\mathcal{O}$ and query them in the coarse-to-fine order to avoid redundant feature evaluations at positions in the empty space.
As depicted in Figure \ref{fig:occ}, when encountering an empty marching point ${\bf x}$, the ray marching process skips that position and advances to the next voxel along the ray direction.
The step size is determined based on the voxel size at the corresponding resolution level where the occupancy grids first indicate that ${\bf x}$ is unoccupied.
Although the coarser-level occupancy grids help to advance with larger step size, the overall ray marching strategy still results in numerous marching points and a significant amount of global memory access to $\mathcal{O}$ at these points.
The impact of frequent global memory access becomes more pronounced, particularly for the rendering process of NGP-RT, where dense computation is not the essential bottleneck.

To further accelerate the rendering process by reducing memory access, we incorporate the occupancy distance grid $\mathcal{G}$ into our ray marching strategy.
Specifically, for each position ${\bf p}=(x,y,z)$, $\mathcal{G}_{\bf p}$ stores the pre-computed distance from ${\bf p}$ to the nearest occupied position of ${\bf p}$. 
To minimize the storage cost, we store the distance as a rounded-down integer in units of the voxel size of $\mathcal{G}$.
With the occupancy distance grid, the step size $s_{\bf p}$ at position ${\bf p}$ in our ray marching strategy is determined as follows:
\begin{equation}
    s_{\bf p} = 
    \begin{cases}
        v \cdot \mathcal{G}_{\bf p}, & \text{if $\mathcal{G}_{\bf p} > 0$}, \\
        s_\mathcal{O}, & \text{otherwise}, \\ 
    \end{cases}
\end{equation}
where $s_\mathcal{O}$ denotes the step size determined by the multi-level occupancy grid, and $v$ denotes the voxel size of the occupancy distance grid.
To reduce overall memory access of $\mathcal{G}$, we only access the distance value when the marching point is unoccupied and the resolution of the exiting occupancy level is smaller than the resolution of $\mathcal{G}$.
As illustrated in Figure \ref{fig:occ}, our ray marching strategy reduces the number of marching points and global memory access by skipping the redundant access to occupancy grids.
In our experiments, we store $\mathcal{G}$ as a $256^3$ grid and save the rounded distance values in the format of {\UrlFont{uint8}}.


\section{Experiments}

We evaluate our method on all 9 scenes of the challenging Mip-NeRF 360 dataset \cite{mipnerf360}, including 5 outdoor scenes and 4 indoor scenes. We refer readers to the supplementary material for implementation details.

\begin{table}[htbp]
    \scriptsize
    \centering
    \caption{Quantitative evaluation of our method compared to previous work on the Mip-NeRF 360 dataset. The upper part presents the results of \textit{offline} rendering methods, while the lower part presents the performance of \textit{real-time} rendering methods.}
    \begingroup
    \setlength{\tabcolsep}{0.5pt}
    \begin{tabular}{ l |c c c | c c c | c c c c}
        \specialrule{1.5pt}{1pt}{1pt}

         \multirow{2}{*}{Method}  &  \multicolumn{3}{c|}{Outdoor Scenes} &  \multicolumn{3}{c|}{Indoor Scenes} &  \multicolumn{4}{c}{All Scenes}  \\
         \cline{2-11}
         & PSNR $\uparrow$ & SSIM $\uparrow$ & LPIPS $\downarrow$   &  PSNR $\uparrow$ & SSIM $\uparrow$ & LPIPS $\downarrow$   &  PSNR $\uparrow$ & SSIM $\uparrow$ & LPIPS $\downarrow$  &  FPS $\uparrow$  \\
        \toprule
        Instant-NGP~\cite{muller2022instant}& 22.80 &  0.569 &  0.365   &  29.15  &  0.870  &  0.221   &  25.62  &  0.703  &  0.301  & 10.4  \\
        Plenoxels~\cite{plenoxels}& 21.68 &  0.513 &  0.491   &  24.83  &  0.766  &  0.426   &  23.08  &  0.625  &  0.301  & 6.79   \\
        Mip-NeRF 360~\cite{mipnerf360} &  24.47 & 0.691 & 0.283   & 31.72 & 0.917 & 0.180  & 27.69  & 0.791 & 0.237 & 0.06 \\
        Zip-NeRF~\cite{barron2023zipnerf} &  \textbf{25.57} & \textbf{0.750} & \textbf{0.207}   & \textbf{32.25} & \textbf{0.926} & \textbf{0.168}  & \textbf{28.54}  & \textbf{0.828} & \textbf{0.189} & 0.49 \\
        \toprule

        \textcolor{lightgray}{Gaussian-7K~\cite{gaussian}}& \textcolor{lightgray}{23.48} & \textcolor{lightgray}{0.654} & \textcolor{lightgray}{0.353} & \textcolor{lightgray}{28.96} & \textcolor{lightgray}{0.907} & \textcolor{lightgray}{0.207} & \textcolor{lightgray}{25.91} & \textcolor{lightgray}{0.766} & \textcolor{lightgray}{0.288} & \textcolor{lightgray}{107} \\
        \cline{1-11}
        Mobile-NeRF~\cite{mobilenerf}& 21.95 & 0.470 & 0.470  & - & - & - & - & - & - & - \\
        BakedSDF~\cite{bakedsdf}& 22.47 & 0.585 & 0.349 & 27.06 & 0.836 & 0.258  & 24.51 & 0.697 & 0.309 & \textgreater 60 \\
        MERF~\cite{reiser2023merf}& \textbf{23.19} & \textbf{0.616} & \textbf{0.343}  & 27.80 & 0.855 & 0.271  & 25.24 & 0.722 & 0.311 & 119\\
        NGP-RT (Ours)  & 22.76 & 0.614 & 0.383  & \textbf{29.25} & \textbf{0.891} & \textbf{0.195}  & \textbf{25.64} & \textbf{0.737} & \textbf{0.299} & 108  \\
        
        \specialrule{1.5pt}{1pt}{1pt}
    \end{tabular}
    \endgroup
    
    \label{tab:accuracy}
\end{table}

\noindent \textbf{Baselines.} We primarily compare our NGP-RT model to two other methods: Instant-NGP, which utilizes implicit multi-level hash features, and MERF, known for its impressive performance with the deferred NeRF architecture. This allows us to assess the effectiveness of integrating NGP-style features into the deferred NeRF architecture.
Additionally, we compare NGP-RT to NeRF-based real-time rendering methods like Mobile-NeRF and BakedSDF. We also include comparisons to offline rendering approaches based on NeRFs like Zip-NeRF, Mip-NeRF 360 and Plenoxels.
In contrast to NeRF methods, 3D Gaussian Splatting \cite{gaussian} has gained popularity in novel view rendering. In this study, we further compare NGP-RT to 3D Gaussians at 7k iterations (Gaussian-7K), which exhibits comparable parameter number and model size to NGP-RT. This comparison highlights the improvements our method brings to NeRF-based approaches.



\subsection{Comparisons}

\begin{figure*}[htbp]
  \centering 
  \includegraphics[width=1.0\linewidth]{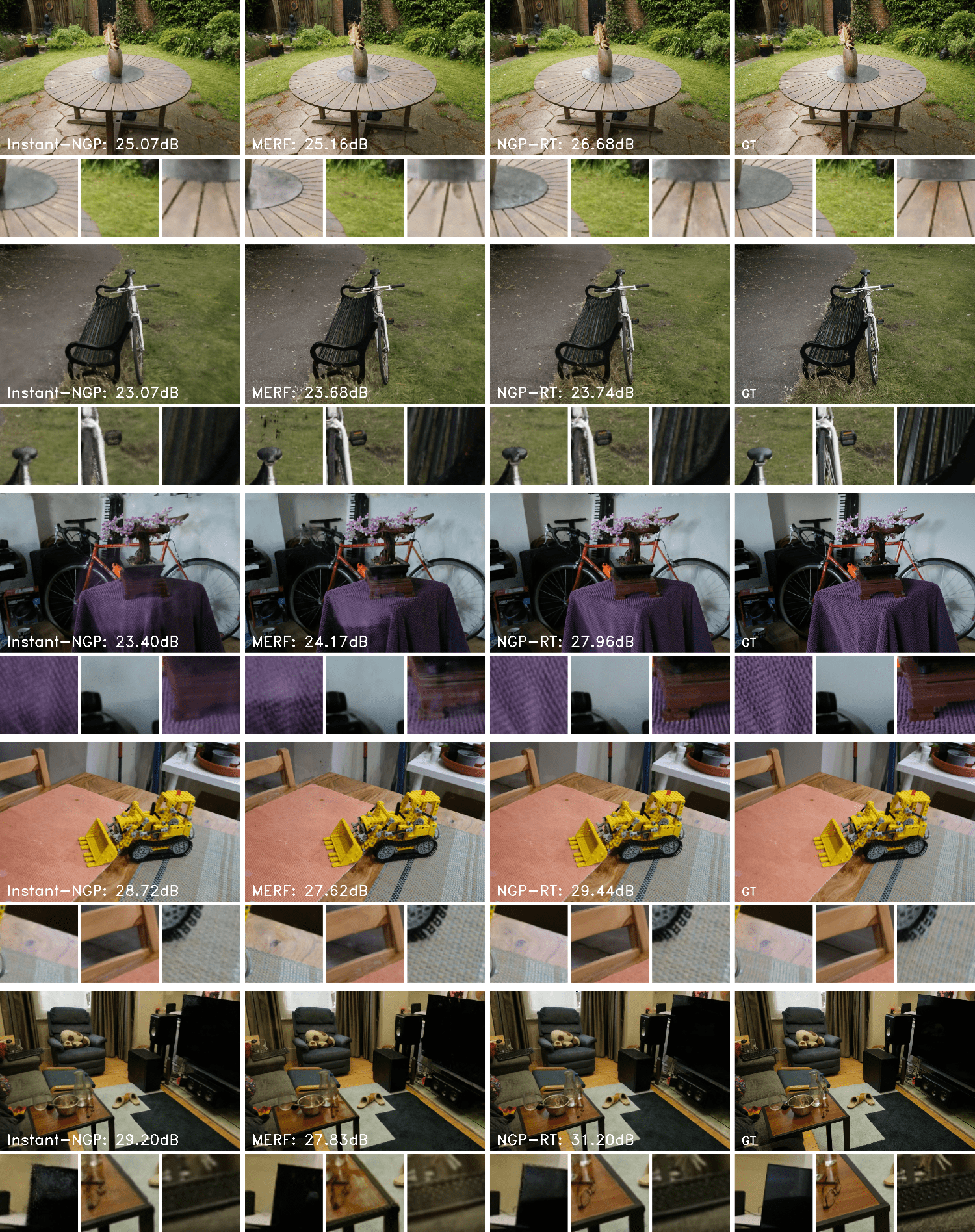}
  \caption{%
    We show comparisons of NGP-RT to previous methods and the ground truth images from several scenes in the Mip-NeRF 360 dataset. NGP-RT avoids inaccurate floaters and presents better light effects in its renderings.  
    %
  }
  \label{fig:comparisons}
\end{figure*}

We evaluate all the above methods regarding the rendering quality and speed at the resolution 1080$\times$1920. We present the quantitative results of PSNR, SSIM, LPIPS and FPS in Table \ref{tab:accuracy} and some qualitative comparisons in Figure \ref{fig:comparisons}.

As shown in Table \ref{tab:accuracy}, NGP-RT brings improvements to the NeRF-based real-time rendering methods (i.e., those methods above the last divider), and achieves better rendering quality on the challenging Mip-NeRF 360 dataset with over 100 fps at 1080p resolution. 
Specifically, NGP-RT achieves comparable rendering quality to Instant-NGP with 10$\times$ rendering speed, demonstrating the effectiveness and efficiency of our lightweight attention mechanism. 
Compared to MERF, NGP-RT achieves similar rendering speed and better overall rendering quality, especially on the indoor scenes.
Although both MERF and NGP-RT utilize a tiny MLP to model the view-dependent appearance, the multi-level hash features employed by NGP-RT have stronger expressing power than the tri-plane features used in MERF.
As shown in Figure \ref{fig:comparisons}, such superior expressing power effectively compensate for the tiny MLP to produce more accurate view-dependent lighting effects, which contributes to the better rendering quality on the indoor scenes.
We also notice that NGP-RT performs slightly worse than MERF on the outdoor scenes, potentially due to the high-frequency subtle structures contained in outdoor scenes that our model cannot model well.

Compared to Gaussian-7K \cite{gaussian}, NGP-RT achieves comparable rendering speed and quality with similar model size, showcasing the effectiveness of the proposed techniques in enhancing NeRF-based methods. Our advancements allow NeRF-based methods to rival the performance of 3D Gaussian Splatting in real-time novel view rendering at 1080P resolution.

\subsection{Ablation Study}

\noindent \textbf{Number of Fine-grained Levels.} 
Since NGP-RT employs the multi-level hash features, we first study the influence of the number of fine-grained hash levels ($L$) that are aggregated by our lightweight attention.

\begin{table}[tbp]
    \scriptsize
    \centering
    
    \caption{Performance of NGP-RT with varying fine-grained hash feature levels $L$.}
    
    \begingroup
    \setlength{\tabcolsep}{7.5pt}
    \begin{tabular}{ l |c c c c c}
        \specialrule{1.5pt}{1pt}{1pt}

         Method  & PSNR $\uparrow$ & SSIM $\uparrow$ & LPIPS $\downarrow$  &  Time(ms) $\downarrow$ & FPS $\uparrow$ \\
        \toprule
        NGP-RT ($L$=2)  & 25.64 & 0.737 & 0.299 & \textbf{9.26} & \textbf{108}  \\
        NGP-RT ($L$=3)  & 25.93 & 0.749 & 0.283 & 12.55 & 79.7\\
        NGP-RT ($L$=4)  & \textbf{26.05} & \textbf{0.753} & \textbf{0.280} & 16.17 & 61.8\\
        \specialrule{1.5pt}{1pt}{1pt}
    \end{tabular}
    \endgroup
    
    \label{tab:levels}
\end{table}


\begin{figure}[tbp]
  \centering 
  \includegraphics[width=0.9\linewidth]{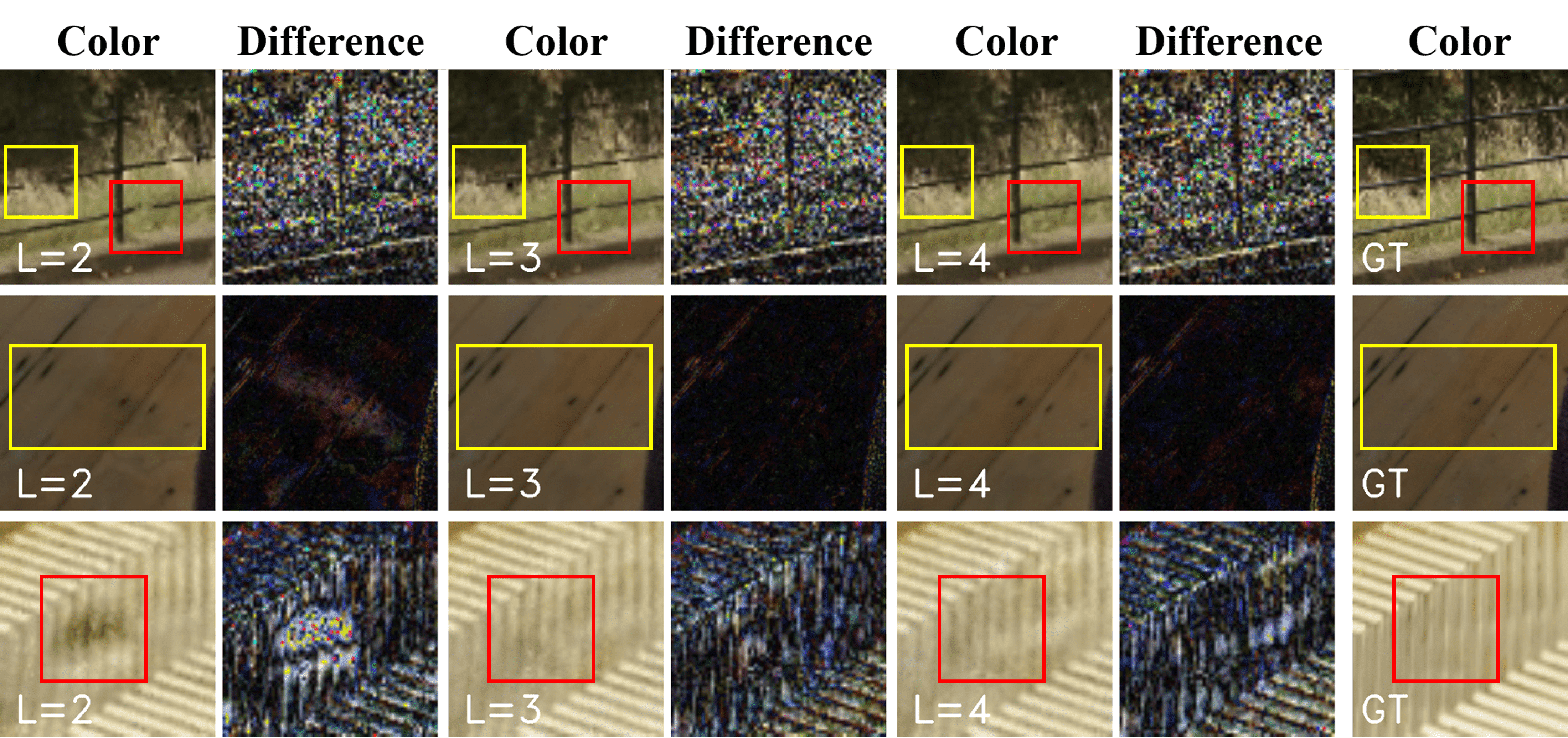}
  \caption{%
    Rendering results from NGP-RT with different fine-grained hash feature levels.
  }
  \label{fig:level}
  \vspace{-2mm}
\end{figure}

As shown in Table \ref{tab:levels} and Figure \ref{fig:level}, our method performs better using more fine-grained feature levels due to their stronger expressing capability for detailed structures and special lighting effects.
On the other hand, the increasing feature levels compromise the rendering speed of NGP-RT, since the global memory access almost dominates the time consumption of NGP-RT. 
In particular, the memory access pattern of NGP-RT is relatively stochastic and leads to limited utilization efficiency of the memory cache, thus greatly influencing the overall rendering speed.
We also observe that NGP-RT ($L=4$) performs slightly better than NGP-RT ($L=3$), indicating that additional hash levels bring limited quality improvements to NGP-RT. 

\noindent \textbf{Feature Aggregation Methods.} 
Based on NGP-RT ($L$=4), we ablate different aggregation methods for the multi-level hash features, including simple sum operation (SUM), tiny MLP (MLP) utilized in Instant-NGP and two modes of lightweight attention (Shared-Att and Separate-Att) with spatially variant/invariant (V/Inv) learnable parameters.
We denote the attention method with a single shared attention parameter for each resolution level with Shared-Att and the attention method with two separate parameters for density and color features of each level with Separate-Att. 
Results summarized in Table \ref{tab:aggregation} demonstrate that Separate-Att (V) achieves comparable rendering quality to the MLP aggregation with more than 4$\times$ FPS.
Compared to the SUM aggregation utilized in MERF and other alternatives of lightweight attention mechanisms, our design achieves the best rendering quality with comparable rendering speed.

\begin{table}[tbp]
    \scriptsize
    \centering
    
    \caption{Performance of NGP-RT ($L$=4) with different feature aggregation methods. We denote the spatially variant/invarant attention parameters with V/Inv.}
    
    
    \begingroup
    \setlength{\tabcolsep}{10pt}
    \begin{tabular}{ l |c c c c c }
        \specialrule{1.5pt}{1pt}{1pt}

         Method  & PSNR $\uparrow$ & SSIM $\uparrow$ & LPIPS $\downarrow$  &  Time $\downarrow$ & FPS $\uparrow$ \\
         \toprule
         MLP  &   26.22 & 0.764 & 0.268 & 68.02 & 14.7\\
        
        \toprule
        SUM  &  25.51 & 0.719 & 0.315 & \textbf{14.94} & \textbf{66.9}  \\
        
        Shared-Att (Inv)  &  25.69 & 0.738 & 0.295 & 15.18 & 65.9 \\
        Separate-Att (Inv)  &  25.73 & 0.740 & 0.291 & 15.13 & 66.1\\
        Shared-Att (V)  & 25.89 & 0.745 & 0.289 & 16.04 & 62.3 \\
        Separate-Att (V)  & \textbf{26.05} & \textbf{0.753} & \textbf{0.280} & 16.17 & 61.8  \\
        \specialrule{1.5pt}{1pt}{1pt}
    \end{tabular}
    \endgroup
    
    \label{tab:aggregation}

    \vspace{-2mm}
\end{table}

\begin{figure}[tbp]
  \centering 
  \includegraphics[width=0.9\linewidth]{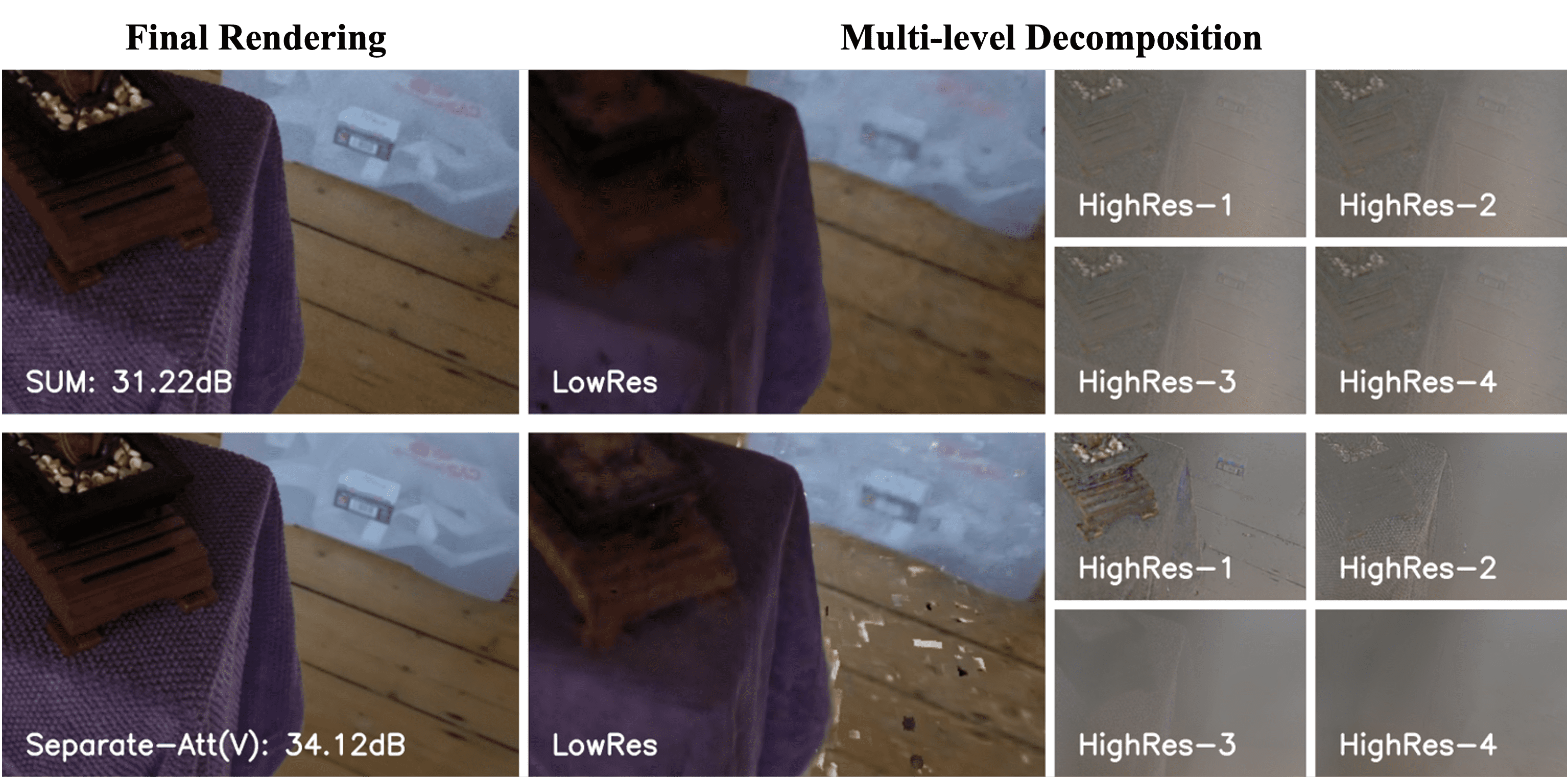}
  \caption{%
    Visualization of contributions of different hash levels to the final renderings. We compare the decomposition of renderings from the SUM and Separate-Att(V) aggregation methods. The gray colors in decomposition results represent zero values. }
  \label{fig:attention}
  \vspace{-2mm}
\end{figure}

We also visualize the contributions of different fine-grained hash levels to the final renderings in Figure \ref{fig:attention}. When visualizing the contribution of a particular feature level, we set the color-related features of all other hash levels to zero values, including the coarse-level features, and render the results with the original multi-level fused densities. 
As shown in Figure \ref{fig:attention}, our lightweight attention mode efficiently splits the detailed textures into different fine-grained hash levels.
The hash collision is greatly alleviated with attention parameters focusing on selected regions instead of the whole scene space.
When multiple 3D positions are mapped to the same entry point of a hash table, a large portion is associated with small attention values, thus avoiding the interference of backward gradient values.

\noindent \textbf{Ray Marching with Occupancy Distance.} 
We conduct experiments on NGP-RT with different fine-grained hash levels to validate the acceleration brought by ray marching with the help of occupancy distance grid $\mathcal{G}$.
As demonstrated by results presented in Table \ref{tab:occ_distance}, the incorporation of $\mathcal{G}$ results in 7\%-10\% acceleration on the rendering speed, which is important for real-time rendering applications.
Additionally, Table \ref{tab:occ_distance} provides the average number of marching points (\# Marching) and the average number of occupied points (\# Occupied) on each ray. Our approach successfully reduces over 40\% of redundant marching points while maintaining the number of valid occupied sample points.

\begin{table}[tbp]
    \scriptsize
    \centering
    
    \caption{NGP-RT rendering speed with or without the occupancy distance grid $\mathcal{G}$.}
    \vspace{-2mm}
    
    \begingroup
    \setlength{\tabcolsep}{9pt}
    \begin{tabular}{ l | c | c | c c c }
        \specialrule{1.5pt}{1pt}{1pt}

         $L$  & $\mathcal{G}$ & PSNR $\uparrow$ & \# Marching $\downarrow$ & \# Occupied $\downarrow$  &  Time (ms) $\downarrow$ \\
        \toprule
        \multirow{2}{*}{2} & w/o & \textbf{25.67} & 85.1 & 17.3 & 9.98 \\
        & w/  &  25.64 & \textbf{46.7} & 17.3 & \textbf{9.26} \\
        \toprule
        \multirow{2}{*}{3} & w/o & 25.91 & 86.4 & 17.6 & 13.89  \\
         & w/  & \textbf{25.93} & \textbf{45.3} & 17.6  & \textbf{12.55}  \\
        \toprule
        \multirow{2}{*}{4} & w/o & \textbf{26.06}  & 85.9 & 18.1 &  18.22 \\
         & w/  & 26.05 & \textbf{46.5} &  18.1 & \textbf{16.17}  \\
        \specialrule{1.5pt}{1pt}{1pt}
    \end{tabular}
    \endgroup

    
    \label{tab:occ_distance}
\end{table}

\section{Conclusion}

In this paper, we present NGP-RT, a novel method that enables real-time rendering of high-quality novel views based on multi-level hash features. By leveraging a lightweight attention mechanism in feature aggregation, NGP-RT preserves the strong representation ability of explicit multi-level hash features while significantly reducing the computational cost compared to the shallow MLP of Instant-NGP. Moreover, we introduce a ray marching strategy with the occupancy distance grid to minimize global memory access during occupancy checks, leading to further improvements in rendering speed. Experimental results demonstrate that NGP-RT achieves high-quality 1080p rendering at over 100 fps, making it suitable for interactive virtual reality experiences and immersive applications. Future research includes extending NGP-RT to handle dynamic scenes and achieve further performance improvements.

\vspace{1mm}
\noindent \textbf{Acknowledgements:}
This work was supported by the Natural Science Foundation of China (62332019) and Beijing Natural Science Foundation (L222008).

\bibliographystyle{splncs04}
\bibliography{main}

\newpage

\clearpage
\renewcommand{\thetable}{S\arabic{table}}
\renewcommand{\thefigure}{S\arabic{figure}}
\renewcommand{\thesection}{S\arabic{section}}
\renewcommand{\theequation}{S\arabic{equation}}
\renewcommand{\thefootnote}{\arabic{footnote}}

\makeatletter

\newcommand*{\@rowstyle}{}

\newcommand*{\rowstyle}[1]{
  \gdef\@rowstyle{#1}%
  \@rowstyle\ignorespaces%
}

\setcounter{page}{1}
\setcounter{table}{0}
\setcounter{section}{0}
\setcounter{equation}{0}
\setcounter{figure}{0}

\title{NGP-RT: Fusing Multi-Level Hash Features with Lightweight Attention for Real-Time Novel View Synthesis \\ Supplementary Material} 

\titlerunning{NGP-RT Supp.}

\author{Yubin Hu\inst{1}\textsuperscript{*}\textsuperscript{$\dag$} \and
Xiaoyang Guo\inst{2}\textsuperscript{*} \and
Yang Xiao \inst{3}
\and \\ 
Jingwei Huang \inst{4}
\and  
Yong-Jin Liu \inst{1}\textsuperscript{\Letter}}

\authorrunning{Y. Hu et al.}

\institute{
BNRist, Department of Computer Science and Technology, Tsinghua University
\and
\mbox{Horizon Robotics \qquad \and Huawei Technologies \qquad \and Game AI Center, Tencent Games}}

\maketitle

\section{Per-scene Results}

\begin{table*}[h]
    \scriptsize
    \centering
    \caption{Per-scene comparison between 3D Gaussian Splattng~\cite{gaussian}, MERF~\cite{reiser2023merf} and our method on the Mip-NeRF 360 dataset.}
    \begingroup
    \setlength{\tabcolsep}{0.5pt}
    \begin{tabu}{ l |l |c c c c c| c c c c | c}
        \specialrule{1.5pt}{1pt}{1pt}

         & \multirow{2}{*}{Method}  &  \multicolumn{5}{c|}{Outdoor Scenes} &  \multicolumn{4}{c|}{Indoor Scenes} &  \multirow{2}{*}{Mean}  \\
         \cline{3-11}
         & & bicycle & flower & garden & stump & tree & room & counter & kitchen & bonsai &     \\
        \toprule
        \multirow{4}{*}{PSNR $\uparrow$} & \textcolor{lightgray}{Gaussian-30K~\cite{gaussian}}& \textcolor{lightgray}{25.06} & \textcolor{lightgray}{21.40} & \textcolor{lightgray}{27.33} & \textcolor{lightgray}{26.67} & \textcolor{lightgray}{22.49} & \textcolor{lightgray}{31.67} & \textcolor{lightgray}{29.09} & \textcolor{lightgray}{31.53} & \textcolor{lightgray}{32.31} & \textcolor{lightgray}{27.51} \\
        
        \rowfont{\color{lightgray}} & Gaussian-7K~\cite{gaussian} & 23.39 & 20.32 & 26.05 & 25.57 & 22.06 & 29.53 & 27.26 & 29.23 & 29.80 & 25.91 \\
        
        \cline{2-12}
        & MERF~\cite{reiser2023merf} & \textbf{22.62} & \textbf{20.33} & 25.58 & \textbf{25.04} & \textbf{22.39} & 29.28 & 25.82 & 27.42 & 28.68 & 25.24  \\
        & NGP-RT (Ours) & 22.40 & 18.88 & \textbf{26.22} & 24.54 & 21.74 & \textbf{29.75} & \textbf{27.27} & \textbf{28.95} & \textbf{31.04} & \textbf{25.64}   \\
        \toprule
        
        \multirow{4}{*}{SSIM $\uparrow$} & \textcolor{lightgray}{Gaussian-30K~\cite{gaussian}}& \textcolor{lightgray}{0.750} & \textcolor{lightgray}{0.590} & \textcolor{lightgray}{0.860} & \textcolor{lightgray}{0.770} & \textcolor{lightgray}{0.640} & \textcolor{lightgray}{0.927} & \textcolor{lightgray}{0.915} & \textcolor{lightgray}{0.932} & \textcolor{lightgray}{0.947} & \textcolor{lightgray}{0.813} \\
        \rowfont{\color{lightgray}} & Gaussian-7K~\cite{gaussian} & 0.640 & 0.510 & 0.810 & 0.720 & 0.590 & 0.903 & 0.887 & 0.910 & 0.928 & 0.766 \\
        \cline{2-12}
        & MERF~\cite{reiser2023merf} & 0.595 & \textbf{0.492} & 0.763 & 0.677 & \textbf{0.554} & 0.874 & 0.819 & 0.842 & 0.884  & 0.722 \\
        & NGP-RT (Ours) & \textbf{0.610} & 0.430 & \textbf{0.820} & \textbf{0.690} & 0.530 & \textbf{0.894} & \textbf{0.852} & \textbf{0.889} & \textbf{0.930} & \textbf{0.737}   \\
        \toprule

        \multirow{4}{*}{LPIPS $\downarrow$} & \textcolor{lightgray}{Gaussian-30K~\cite{gaussian}}& \textcolor{lightgray}{0.244} & \textcolor{lightgray}{0.360} & \textcolor{lightgray}{0.122} & \textcolor{lightgray}{0.243} & \textcolor{lightgray}{0.347} & \textcolor{lightgray}{0.197} & \textcolor{lightgray}{0.183} & \textcolor{lightgray}{0.116} & \textcolor{lightgray}{0.180} & \textcolor{lightgray}{0.211} \\
        \rowfont{\color{lightgray}} & Gaussian-7K~\cite{gaussian} & 0.374 & 0.440 & 0.186 & 0.328 & 0.435 & 0.239 & 0.229 & 0.149 & 0.212 & 0.288 \\
        \cline{2-12}
        & MERF~\cite{reiser2023merf} & \textbf{0.371} & \textbf{0.406} & 0.215 & \textbf{0.309} & \textbf{0.414} & 0.292 & 0.307 & 0.224 & 0.262  & 0.311 \\
        & NGP-RT (Ours) & 0.400 & 0.485 & \textbf{0.214} &  0.366 & 0.449 & \textbf{0.197} & \textbf{0.231} & \textbf{0.165} & \textbf{0.186} & \textbf{0.299}   \\
        \toprule

        \multirow{4}{*}{Time(ms) $\downarrow$} & \textcolor{lightgray}{Gaussian-30K~\cite{gaussian}}& \textcolor{lightgray}{19.65} & \textcolor{lightgray}{10.02} & \textcolor{lightgray}{16.67} & \textcolor{lightgray}{11.54} & \textcolor{lightgray}{11.21} & \textcolor{lightgray}{9.65} & \textcolor{lightgray}{9.12} & \textcolor{lightgray}{10.99} & \textcolor{lightgray}{7.30} & \textcolor{lightgray}{11.79} \\
        \rowfont{\color{lightgray}} & Gaussian-7K~\cite{gaussian} & 12.70 & 7.62 & 13.40 & 9.17 & 8.15 & 8.03 & 8.43 & 10.25 & 6.34 & 9.34  \\
        \cline{2-12}
        
        
        & MERF~\cite{reiser2023merf} & - &  - &  -   &  -  &  -  &  -   &  -  &  -  &  -  & \textbf{8.40} \\
        & NGP-RT (Ours) & 9.00 & 11.70 & 8.72 & 10.23 & 10.29 & 7.97 & 10.15 & 7.37 & 7.85 & 9.25   \\

        
        \specialrule{1.5pt}{1pt}{1pt}
    \end{tabu}
    \endgroup
    
    \label{tab:per-scene-result}
\end{table*}

We present a comparison of our method NGP-RT ($L=2$, $L_C=512$) with two recently prominent techniques, 3D Gaussian Splatting~\cite{gaussian} and MERF~\cite{reiser2023merf}, for real-time rendering of unbounded 360-degree scenes. 
The per-scene results are summarized in Table~\ref{tab:per-scene-result}, including metrics such as PSNR, SSIM, LPIPS for rendering quality, and rendering time in \textit{ms} at a resolution of 1080$\times$1920.

\setcounter{footnote}{0}

Regarding the rendering time of MERF, we only provide the average time across all views calculated from the fps number reported in the original paper, as the released jax code\footnote{https://github.com/google-research/google-research/tree/master/merf} does not support real-time rendering of the baked MERF model. With the provided real-time rendering solution using web viewer, we cannot accurately measure the rendering time from the test views due to the upper bound of the rendering speed.

As for the rendering time of 3D Gaussian Splatting and NGP-RT, we measure them on the same Nvidia RTX 3090 GPU to ensure fairness.
We evaluate 3D Gaussian Splatting with two configurations, 7K and 30K iterations, following the original paper.
The Gaussian-30K version achieves better rendering quality with larger number of 3D gaussians, which leads to the slower rendering speed.
As for the Gaussian-7K version, NGP-RT produces comparable rendering quality to it with similar rendering speed. 

Different from our 1080$\times$1920 resolution for evaluation, the original paper of 3D GS evaluates at the ``reference resolution'', which is specified as the resolution of \textit{images\_2} for indoor scenes (${\sim}$1000$\times$1500) and the resolution of \textit{images\_4} for outdoor scenes (${\sim}$800$\times$1200).
To perform a more comprehensive comparison, we also evaluate NGP-RT at the same ``reference resolution'' and present the results in Table~\ref{tab:per-scene-result-reference}.
The experimental results of 3D GS are directly copied from the original paper.

As demonstrated by the above comparisons, NGP-RT produces comparable rendering quality to Gaussian-7K with similar rendering speed, showing that our advancements allow NeRF-based methods to rival the performance of 3D Gaussian Splatting with similar model size (250MB for NGP-RT v.s. 523MB for Gaussian-7K).

\section{Implementation Details}
\label{sec:rationale}

\subsection{Network Design}

\textbf{Multi-level Hash Features.} In NGP-RT, we set the resolution of the coarse-grained level to $L_C = 512$, and the resolutions of the fine-grained levels to 1024 and 2048 for $L=2$.
With each increase in $L$, we double the resolution of the finest level.
As mentioned in Section~\ref{sec: overview}, we employ an auxiliary NGP model to optimize the coarse-grained features and attention parameters. 
This auxiliary model comprises 6 coarse-grained resolution levels ranging from 16 to $L_C=512$. Each coarse-grained level has a hash table with a maximum length of $2^{21}$ and a feature dimension of 4. After training and baking the features into feature grids, this auxiliary NGP model will be discarded.
Regarding the $L$ fine-grained resolution levels, we set the length of the hash tables to $2^{22}$ to accommodate diverse high-resolution features.
The hash tables at the fine-grained levels consist of explicit 8-dimensional deferred NeRF features, as opposed to the implicit features used in the coarse-grained levels.

\textbf{Forward Pass.} For 3D points sampled along the casting ray of a training pixel, we extract coarse-grained hash features from different resolution levels. These features are then concatenated into a 24-dimensional vector. We feed this vector into a shallow MLP with 1 hidden layer containing 64 neurons.
Subsequently, the MLP decodes the concatenated features into a $(8+2L)$-dimensional coarse-grained feature. This feature includes a 1-dimensional density value $\Tilde{\sigma}$, a 3-dimensional RGB color value $\Tilde{\bf c}_{\bf d}$, a 4-dimensional feature vector $\Tilde{\bf f}_{\bf s}$ for view-dependent colors, and a $2L$-dimensional set of attention parameters ${\bf a} = [\omega^{1}, \beta^{1}, \dots, \omega^{L}, \beta^{L}]$ for the $L$ fine-grained resolution levels.
The attention parameters are used to aggregate the fine-grained hash features, while the density values are employed to accumulate the color features through volume rendering. After the alpha composition process, we utilize another MLP, denoted as ${\rm MLP}\psi$, to decode the view-dependent colors according to Eq.~({\color{red}5}). ${\rm MLP}\psi$ consists of two hidden layers, each with 64 neurons.
In addition to the accumulated view-dependent features, ${\rm MLP}\psi$ takes the embedding of the view direction as input. This view direction embedding is represented by a 16-dimensional vector that encodes a 4-degree spherical harmonics representation.

\begin{table*}[th]
    \scriptsize
    \centering
    \caption{Per-scene comparisons between 3D Gaussian Splattng~\cite{gaussian} and our method using the resolution settings utilized in the 3D Gaussian Splatting paper on the Mip-NeRF 360 dataset.}
    \begingroup
    \setlength{\tabcolsep}{1.5pt}
    \begin{tabu}{ l |l |c c c c c| c c c c | c}
        \specialrule{1.5pt}{1pt}{1pt}

         & \multirow{2}{*}{Method}  &  \multicolumn{5}{c|}{Outdoor Scenes} &  \multicolumn{4}{c|}{Indoor Scenes} &  \multirow{2}{*}{Mean}  \\
         \cline{3-11}
         & & bicycle & flower & garden & stump & tree & room & counter & kitchen & bonsai &     \\
        \toprule

        \multirow{3}{*}{PSNR $\uparrow$} & \textcolor{lightgray}{Gaussian-30K~\cite{gaussian}} & \textcolor{lightgray}{25.25} & \textcolor{lightgray}{21.52} & \textcolor{lightgray}{27.41} & \textcolor{lightgray}{26.55} & \textcolor{lightgray}{22.49} & \textcolor{lightgray}{30.63} & \textcolor{lightgray}{28.70} &  \textcolor{lightgray}{30.32} & \textcolor{lightgray}{31.98} & \textcolor{lightgray}{27.21} \\
        
        \rowfont{\color{lightgray}} & Gaussian-7K~\cite{gaussian} & 23.60 & 20.52 & 26.25 & 25.71 & 22.09 & 28.14 & 26.71 & 28.55 & 28.85 & 25.60  \\

        \cline{2-12}
        & NGP-RT (Ours) & 22.50 & 19.03 & 26.19 & 24.70 & 21.68 & 29.81 & 27.24 & 28.94 & 31.03 & 25.68   \\
        \toprule

        \multirow{3}{*}{SSIM $\uparrow$} & \textcolor{lightgray}{Gaussian-30K~\cite{gaussian}} & \textcolor{lightgray}{0.771} & \textcolor{lightgray}{0.605} & \textcolor{lightgray}{0.868} & \textcolor{lightgray}{0.775} & \textcolor{lightgray}{0.638} & \textcolor{lightgray}{0.914} & \textcolor{lightgray}{0.905} & \textcolor{lightgray}{0.922} & \textcolor{lightgray}{0.938} & \textcolor{lightgray}{0.815} \\
        
        \rowfont{\color{lightgray}}  & Gaussian-7K~\cite{gaussian} & 0.675 & 0.525 & 0.836 & 0.728 & 0.598 & 0.884 & 0.873 & 0.900 & 0.910 & 0.770   \\

        \cline{2-12}
        
        & NGP-RT (Ours) & 0.621 & 0.425 & 0.823 & 0.692 & 0.537 & 0.888 & 0.853 & 0.885 & 0.929 & 0.739  \\
        \toprule

        \multirow{3}{*}{LPIPS $\downarrow$}  & \textcolor{lightgray}{Gaussian-30K~\cite{gaussian}} & \textcolor{lightgray}{0.205} & \textcolor{lightgray}{0.336} & \textcolor{lightgray}{0.103} & \textcolor{lightgray}{0.210} & \textcolor{lightgray}{0.317} & \textcolor{lightgray}{0.220} & \textcolor{lightgray}{0.204} & \textcolor{lightgray}{0.129} & \textcolor{lightgray}{0.205} & \textcolor{lightgray}{0.214} \\
        
        \rowfont{\color{lightgray}}  & Gaussian-7K~\cite{gaussian} & 0.318 & 0.417 & 0.153 & 0.287 & 0.404 & 0.272 & 0.254 & 0.161 & 0.244 & 0.279  \\

        \cline{2-12}
        
        & NGP-RT (Ours) & 0.391 & 0.475 & 0.215 &  0.360 & 0.437 & 0.201 & 0.231 & 0.162 & 0.187 & 0.295 \\
        \toprule

        \multirow{3}{*}{FPS $\uparrow$} & \textcolor{lightgray}{Gaussian-30K~\cite{gaussian}} & \textcolor{lightgray}{-} & \textcolor{lightgray}{-} & \textcolor{lightgray}{-} & \textcolor{lightgray}{-} & \textcolor{lightgray}{-} & \textcolor{lightgray}{-} & \textcolor{lightgray}{-} & \textcolor{lightgray}{-} & \textcolor{lightgray}{-}  & \textcolor{lightgray}{134} \\
        
        \rowfont{\color{lightgray}}  & Gaussian-7K~\cite{gaussian} & - & - & - & - & - & - & - & - & -  & 160  \\

        \cline{2-12}
        
        & NGP-RT (Ours) & 175 & 139 & 183 & 161 & 167 & 190 & 171 & 201 & 189 & 175  \\
        
        \specialrule{1.5pt}{1pt}{1pt}
        
    \end{tabu}
    \endgroup
    
    \label{tab:per-scene-result-reference}
\end{table*}

\textbf{Activation Layers.}
In NGP-RT, we employ different activation functions and strategies for features of different modalities. 

1) For the per-point density value $\sigma$, we apply the exponential function with truncated gradients for activation following~\cite{muller2022instant}. 


2) For the per-point and per-level attention parameters $\{\omega^l, \beta^l\}$, we utilize a Sigmoid layer to normalize them to the range of $[0,1]$.

3) As for features of the color modality, we first compute the RGB values for the training pixel using Eq.~({\color{red}5}). Then, we apply a Sigmoid layer to normalize the color components to the range of $[0,1]$.


\textbf{Number of Feature Parameters.}
At the rendering stage, we discard the auxiliary NGP model and bake the coarse-grained features into a sparse feature grid $\Tilde{\mathcal{F}}$.
The dimensions of $\Tilde{\mathcal{F}}$ are $L_C \times L_C \times L_C \times (8+2L)$, where $L_C$ represents the coarse-grained resolution, and $L$ denotes the number of fine-grained resolution levels. 
The sparsity ratio is around 2\%, which is similar to that of MERF.
Overall, the number of feature parameters of the NGP-RT model in Table~{\color{red}1} is comparable to that in MERF, which contains a $512\times512\times512\times8$ coarse-grained feature grid with sparsity ratio of 2\% and three $2048\times2048\times8$ feature planes.
Thereby the comparisons between NGP-RT ($L=2$, $L_C=512$) and MERF in Table~{\color{red}1} and Table~\ref{tab:per-scene-result} are fair. 
We can calculate the number of parameters for both models as follows:
\begin{equation}
\label{eq:calc_numbers}
\begin{split}
    & N_{\rm MERF} = 8\times(0.02\times512^3 + 3\times2048^2)=122.1 {\rm M}, \\
    & N_{\rm ours} = 12 \times 0.02\times L_C^3 + 8 \times L\times2^{22}=99.32 {\rm M}.
\end{split}
\end{equation}

\subsection{Loss Functions}

We train NGP-RT with the Huber Loss~\cite{huber} $\mathcal{L}_{color}$ of predicted color values and the regularizer $\mathcal{L}_{dist}$ proposed in Mip-NeRF 360~\cite{mipnerf360}. 
The overall loss function for each training pixel can be formulated as follows, 
\begin{equation}
\label{eq:calc_numbers}
\begin{split}
    & \mathcal{L} = \mathcal{L}_{color} + \eta \mathcal{L}_{dist}, \\
    & \mathcal{L}_{color} = \mathcal{L}_{Huber}({\bf C}_{pred}, {\bf C}_{gt}), \\
    & \mathcal{L}_{dist} = \sum_{i,j} {\rm w}_i {\rm w}_j |\frac{t_i+t_{i+1}}{2} - \frac{t_j+t_{j+1}}{2}| \\
    & \quad \quad \quad + \frac{1}{3}\sum_i {\rm w}^2_i(t_{i+1}-t_i),
\end{split}
\end{equation}
where ${\rm w}_i$ denotes the rendering weight and $t_i$ denotes the ray distance. 
In our experiments, we set $\eta=0.01$ to adjust the strength of the regularization term and effectively regularize the underlying radiance field of NGP-RT.

We also observe that a high $\eta$ leads to missing parts (e.g., petals and leaves) in the renderings of intricate small structures, resulting in degenerated rendering quality for outdoor scenes. 
As shown in Table \ref{tab:r3}, by tuning the weight $\eta$ for each outdoor scene, we achieve comparable rendering quality to MERF while maintaining an acceptable average rendering speed (102 FPS) across all Mip-NeRF 360 scenes. 

\begin{table}[thbp]
\scriptsize
\centering
\caption{Experimental results with tuned weights $\eta$ for $\mathcal{L}_{dist}$.}
\label{tab:r3}
\begingroup
\setlength{\tabcolsep}{5pt}
\begin{tabular}{ c | c | c | c | c | c | c | c}
\specialrule{1.5pt}{1pt}{1pt}
  & bicycle & flower & garden & stump & treehill & outdoor & Mip-NeRF 360 \\
\toprule
$\eta$ &  0.005 & 0.001 & 0.010 & 0.001 & 0.005 & - & - \\
\hline
PSNR$_{Ours}$ & \textbf{22.71} & 19.86 & \textbf{26.22} & 24.93 & \textbf{22.43} & \textbf{23.23} &  \textbf{25.91} \\
\hline
PSNR$_{MERF}$ & 22.62 & \textbf{20.33} & 25.58 & \textbf{25.04} &  22.39 & 23.19 & 25.24\\

\specialrule{1.5pt}{1pt}{1pt}

\end{tabular}
\endgroup

\vspace{-4.5mm}
\end{table}

\subsection{Auto-tuning of Marching Step Size}

\begin{figure*}[htbp]
  \centering 
  \includegraphics[width=1.0\linewidth]{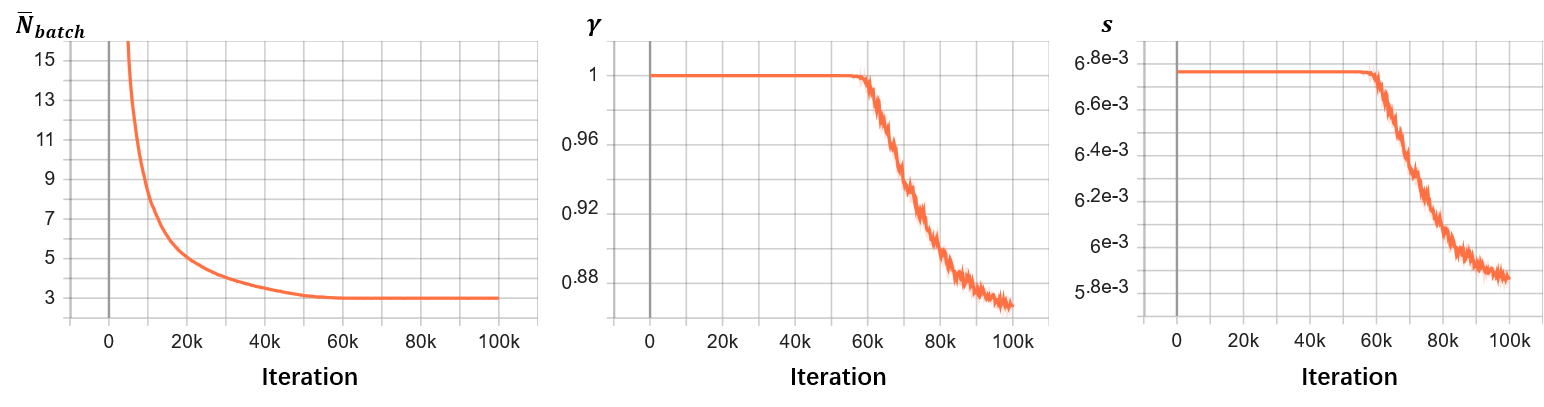}
  \caption{%
    Illustration of the optimization process of $\Bar{N}_{batch}$, $\gamma$ and $s$.
  }
  \label{fig:stepsize}
\end{figure*}

During the training stage of NGP-RT, the regularization term $\mathcal{L}_{dist}$ gradually sparsifies the underlying radiance field of NGP-RT, which results in fewer sampled points along each casting ray.
Such a reduction on the number of sampled points alleviates the burden of frequent feature access and helps to achieve real-time rendering.
However, the strength of regularization can make our model get stuck into suboptimals with only one even no sampled points on the casting rays.
To prevent the collapse into an excessively small number of sample points, we introduce an auto-tuning strategy of ray marching step size when the number of sampled points hits a minimum value $N_{min}$ at the training stage.
Specifically, we count the average number of sampled points $\Bar{N}_{batch}$ across all casting rays inside the training batch.
According to $\Bar{N}_{batch}$ of the last training batch, we design a strategy to automatically tune a scaler $\gamma$ for the marching step size to avoid over sparsification. The auto-tuning strategy can be formulated as follows,
\begin{equation}
\label{eq:autotune}
\begin{split}
    & s = \gamma \cdot s_0, \\
    & \gamma = 
    \begin{cases}
        \min \{1.0, \ (1+\kappa)\cdot \gamma \}, & \text{if $\Bar{N}_{batch} \geq N_{min}$ }, \\ 
        \max \{\gamma_{min}, (1-\kappa)\cdot \gamma \}, & \text{otherwise},
    \end{cases}
\end{split}
\end{equation}
where $s_0$ denotes the base step size varying according to the specially designed contraction function, $s$ denotes the scaled step size which is utilized in ray marching, and $\gamma_{min}$ denotes the minimum value for the scaler $\gamma$.
In our experiments, we set the base step size inside our $\left[-1,1\right]^3$ ROI grid to $\frac{2\sqrt{3}}{512}$, 
$N_{min}=3$ to reserve a small number of marching points, $\kappa=0.001$ for the progressive multiplier, $\gamma_{min}=0.2217$ for the minimum value of scaler $\gamma$, and the initial value of $\gamma$ to 1.0.
We plot the optimization progress of $\Bar{N}_{batch}$, $\gamma$ and $s$ in Figure~\ref{fig:stepsize}, which reflects the effectiveness of our auto-tuning strategy.

\subsection{Optimization}
We train NGP-RT for 100k iterations using an Adam optimizer with an linearly decaying learning rate. 
The learnig rate is warmed up during the first 1k iterations where it is increased from 0 to 0.01, and then decay to 0 in the following iterations.
The weighting parameter $\eta$ for $\mathcal{L}_{dist}$ is linearly warmed up in the first 50k iterations from 0 to 0.01, and then remains as a constant value.
Adam's hyperparameters $\beta_1$, $\beta_2$ and $\epsilon$ are set to 0.9, 0.99 and 1e-8, respectively.
When collecting the training batch of rays, we expect the total number of sample points to be less than $2^{21}$ and the total number of rays to be less than $8000$. 

\subsection{Real-time Rendering}

At the rendering stage, we utilize several techniques to further improve the rendering speed.

1) We employ the visibility culling and occupancy dilation techniques similar to MERF~\cite{reiser2023merf} to generate the occupancy grid for rendering, which contains much less occupied positions compare to that utilized at the training stage. The occupancy grid is then down-sampled to 5 resolution levels including $512^3$, $256^3$, $128^3$, $64^3$, and $32^3$ for fast occupancy check.

2) We stop the ray marching of a casting ray when the transmittance value of a sampled point is less than 2e-3, in order to avoiding redundant samplings in the empty space behind the visible surfaces.

3) We utilize the fully fused MLP implemented in tiny-cuda-cnn\footnote{https://github.com/NVlabs/tiny-cuda-nn} to accelerate the execution of our view-dependent ${\rm MLP}_\psi$.

4) We implement a CUDA-based C++ program to render NGP-RT at the fast speed. Our implementation utilizes aligned memory to increase the efficiency of memory fetching performed by CUDA kernels.

\section{Analysis of Hash Feature Attention}

\subsection{Example of Hash Collision Reduction} 

To demonstrate the collision reduction effects of our attention mechanism, we examine two 3D points in the bonsai scene: Point $P_A$ on the detailed flowers and Point $P_B$ on the flat floor. 

At the 2048 resolution, two corners of their enclosing voxels collide after the hash function. Therefore both points influence the learning of the same hash entry $H$.
As shown in Table \ref{tab:r2}, $P_A$ and $P_B$ have similar influence weight $\xi$ on $H$ due to the similar point-to-corner distance.
Similar $\xi$ values result in hash collision and impede the optimization of $H$ due to averaged gradient directions.

Differently, NGP-RT disambiguates $P_A$ and $P_B$ using different $\omega$ and $\beta$ values. This offloads the influence of  $P_B$ to another resolution level and alleviates the hash collision. As a result, the improved allocation of hash features contributes to a notable enhancement in rendering the bonsai scene.
Please refer to ~\Cref{fig:attention-param-1,fig:attention-param-2,fig:attention-param-3,fig:attention-param-4} for more visualizations of the attention weights allocation. 


\vspace{-6.5mm}

\begin{table}[thbp]
\scriptsize
\centering
\caption{Pre-attention weights $\xi$ and post-attention weights $\xi'$.}
\label{tab:r2}
\begingroup
\setlength{\tabcolsep}{1.5pt}
\begin{tabular}{c | c | c | c | c | c | c }
\specialrule{1.5pt}{1pt}{1pt}
$P_X$ & Resolution & $\xi_X $ & $\omega(X)$ & $\beta(X)$ & $\xi'_{X, \omega} = \xi_X \cdot \omega(X)$ & $\xi'_{X, \beta} = \xi_X \cdot \beta(X)$ \\

\toprule
$P_A$ & 2048 & 0.436 & \textbf{0.257} & \textbf{0.636} & \textbf{0.112} & \textbf{0.277} \\
\hline
$P_B$ & 2048 & \textbf{0.515} & 0.033 & 0.164 & 0.017 & 0.084 \\
\toprule
$P_B$ & 1024 & 0.604 & 0.349 & 0.671 & 0.211 & 0.405 \\

\specialrule{1.5pt}{1pt}{1pt}

\end{tabular}
\endgroup

\vspace{-4.5mm}
\end{table}

\subsection{Visualization of Attention Parameters}

\begin{figure*}[htbp]
  \centering 
  \includegraphics[width=0.7\linewidth]{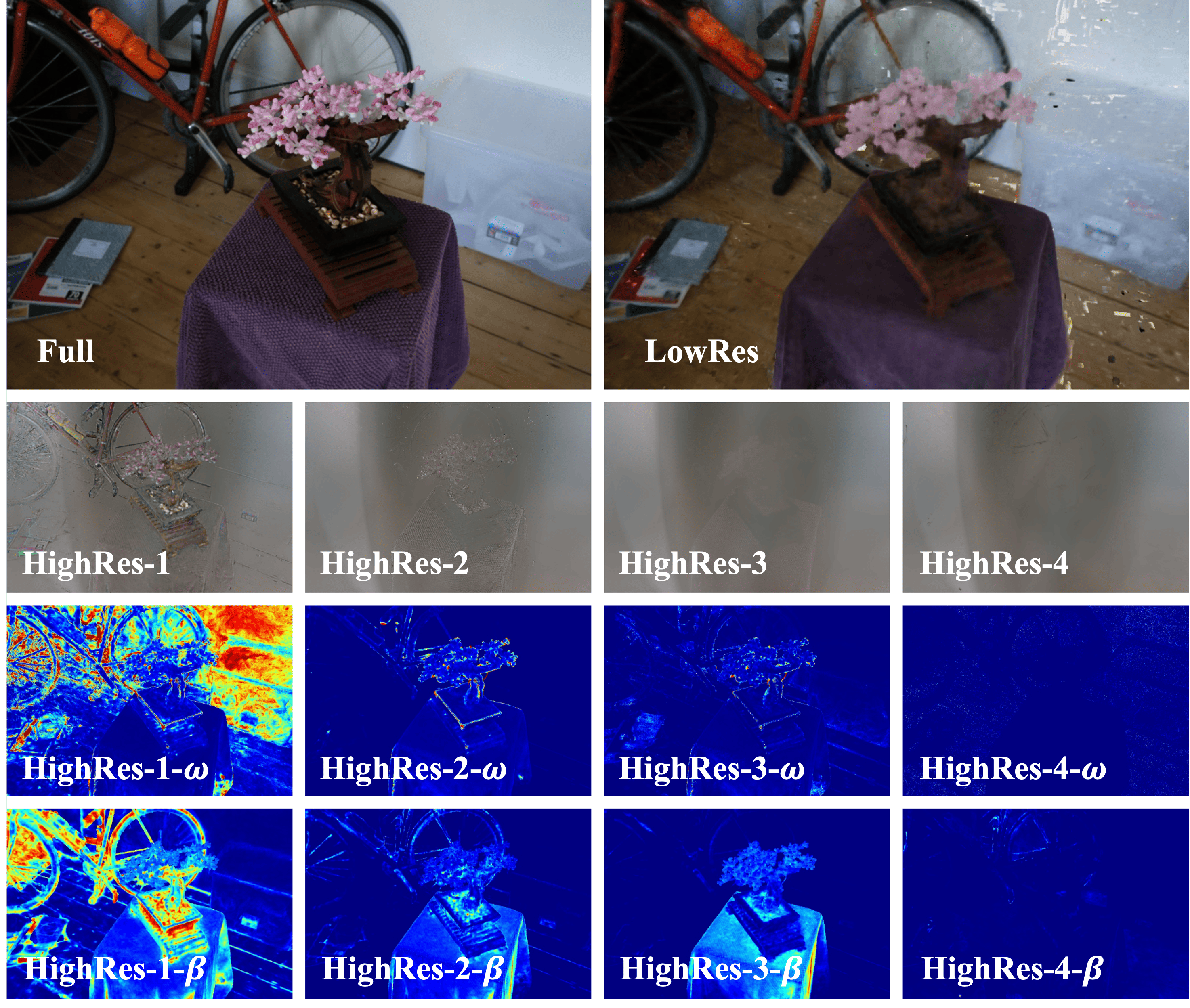}
  \caption{%
    Visualization of the color decomposition and multi-level attention parameters for the density values and color-related features.
  }
  \label{fig:attention-param-1}
\end{figure*}

\begin{figure*}[htbp]
  \centering 
  \includegraphics[width=0.7\linewidth]{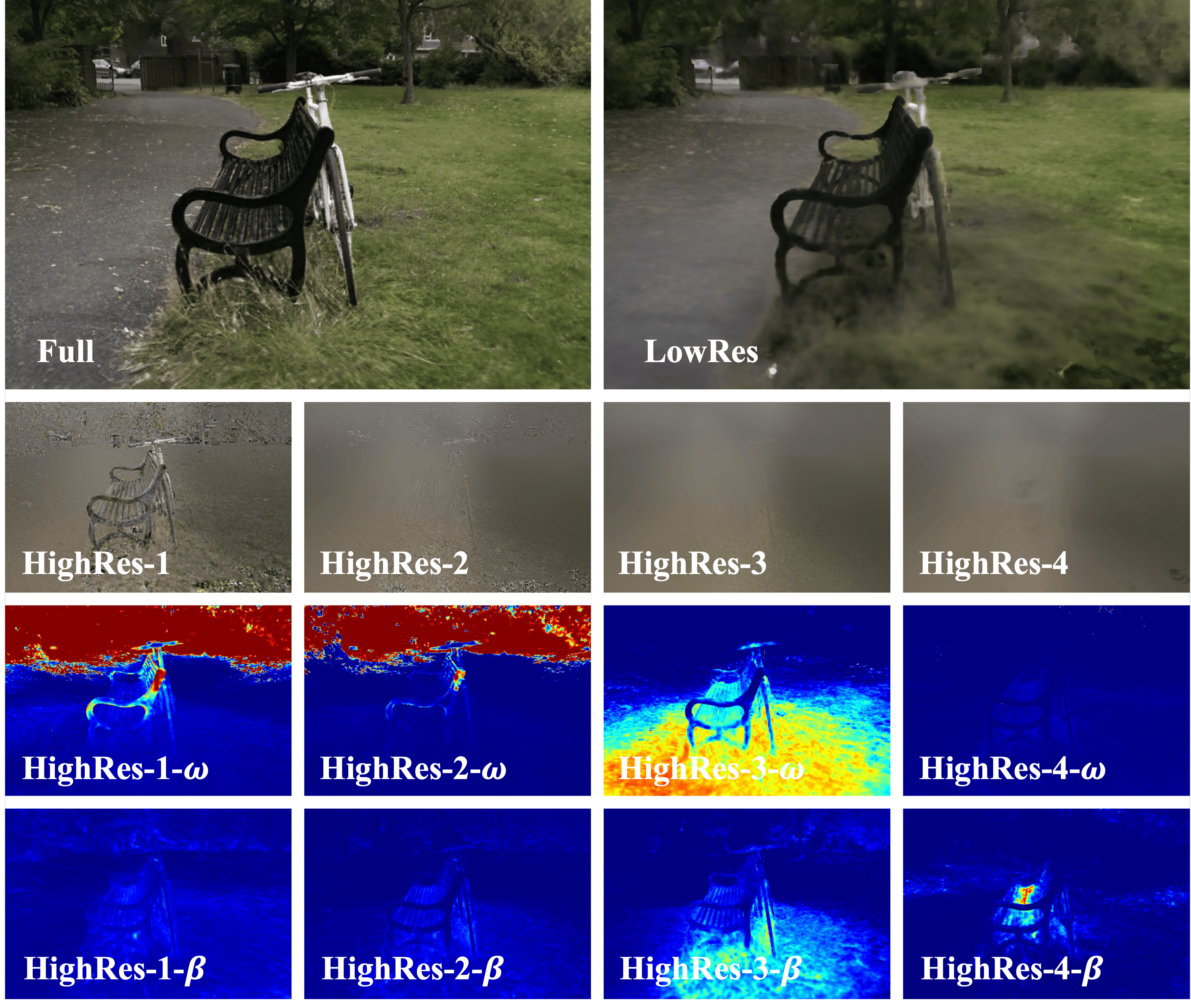}
  \caption{%
    Visualization of the color decomposition and multi-level attention parameters for the density values and color-related features.
  }
  \label{fig:attention-param-2}
\end{figure*}

\begin{figure*}[htbp]
  \centering 
  \includegraphics[width=0.7\linewidth]{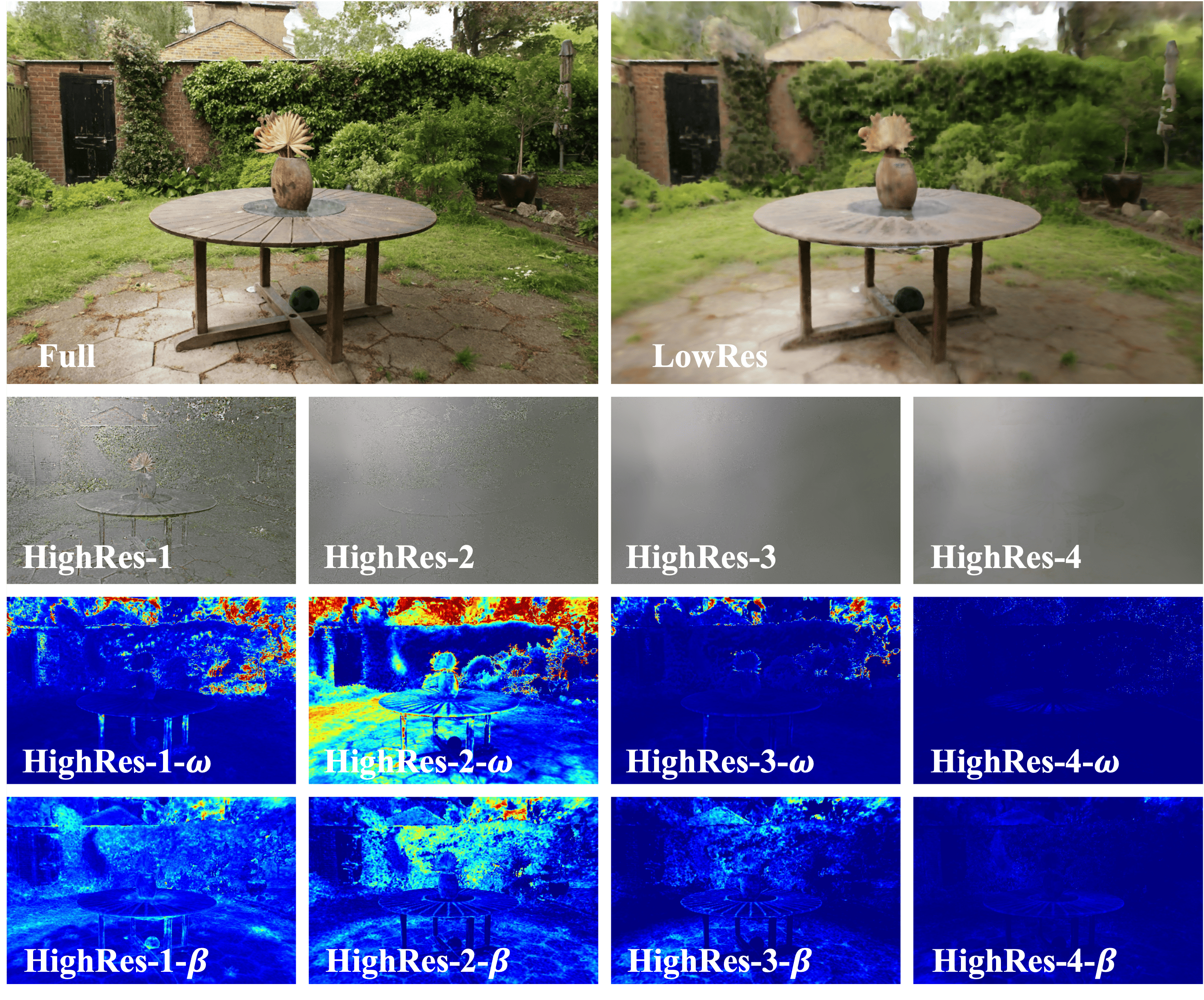}
  \caption{%
    Visualization of the color decomposition and multi-level attention parameters for the density values and color-related features.
  }
  \label{fig:attention-param-3}
\end{figure*}

\begin{figure*}[htbp]
  \centering 
  \includegraphics[width=0.7\linewidth]{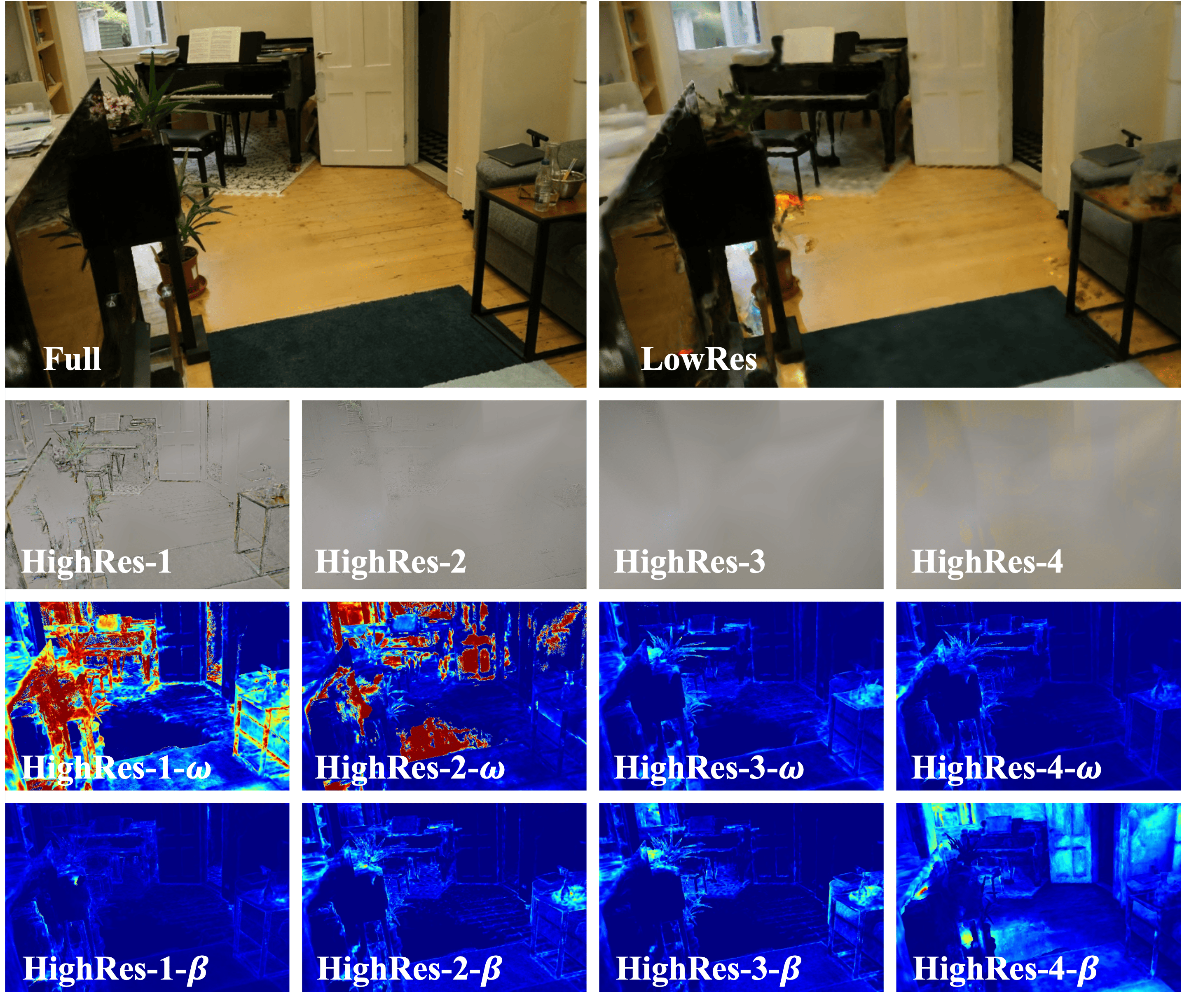}
  \caption{%
    Visualization of the color decomposition and multi-level attention parameters for the density values and color-related features.
  }
  \label{fig:attention-param-4}
\end{figure*}

Besides the visualizations of multi-level color contributions in Figure~\ref{fig:attention}, we further visualize the post-Sigmoid attention parameters of NGP-RT ($L=4$) in~\Cref{fig:attention-param-1,fig:attention-param-2,fig:attention-param-3,fig:attention-param-4} to illustrate the functionality of the proposed lightweight attention mechanism.
We generate the visualization results by applying volume rendering to the attention weights. 

As shown in the visualizations, different fine-grained levels focus on different regions with the help of our lightweight attention mechanism, which sufficiently exploits the expressive power of multi-level hash features.
We also notice that the regions-of-interest for attention parameters $\omega$ and $\beta$ are different, indicating that it is necessary to employ separate attention parameters for the density values and color-related features.

\end{document}